\RequirePackage{fix-cm}
\documentclass[twocolumn]{svjour3}          % twocolumn
\smartqed  % flush right qed marks, e.g. at end of proof
\usepackage{graphicx}
\usepackage[pagebackref=true,breaklinks=true,letterpaper=true,colorlinks,bookmarks=false,citecolor=blue,linkcolor=blue]{hyperref}
\usepackage{amsmath,amssymb}
\usepackage{url}
\usepackage{paralist}
\usepackage{subfig}
\usepackage{booktabs}
\usepackage{makecell}
\usepackage{gensymb}
\usepackage{pifont}
\newcommand{\cmark}{\ding{51}}%
\usepackage{xspace}
\usepackage{color}
\usepackage{multirow}
\usepackage{ifthen}

\newcommand{\tb}[1]{\textbf{#1}}

\long\def\ignorethis#1{}

\definecolor{gray}{rgb}{0.35,0.35,0.35}
\definecolor{MyBlue}{rgb}{0,0.2,0.8}
\definecolor{MyRed}{rgb}{0.8,0.2,0}
\definecolor{MyGreen}{rgb}{0.0,0.5,0.1}
\definecolor{MyGray}{rgb}{0.4,0.4,0.4}

\newlength\paramargin
\newlength\figmargin
\newlength\subfigmargin
\newlength\secmargin
\newlength\subsecmargin
\newlength\tabmargin
\newlength\eqmargin

\usepackage{array}
\newcolumntype{L}[1]{>{\raggedright\let\newline\\\arraybackslash\hspace{0pt}}m{#1}}
\newcolumntype{C}[1]{>{\centering\let\newline\\\arraybackslash\hspace{0pt}}m{#1}}
\newcolumntype{R}[1]{>{\raggedleft\let\newline\\\arraybackslash\hspace{0pt}}m{#1}}

\newcommand{\mpage}[2]
{
\begin{minipage}[t]{#1\linewidth}\centering
#2
\end{minipage}
}

\newcommand{\Paragraph}[1]
{\vspace{2mm} \noindent \textbf{#1}}

\def\ie{i.e.,~}
\def\eg{e.g.,~}

\def\vs{vs.~}

\newcommand{\subsecref}[1]{Section~\ref{subsec:#1}}
\newcommand{\figref}[1]{Figure~\ref{figure:#1}}
\newcommand{\tabref}[1]{Table~\ref{tab:#1}}
\newcommand{\eqnref}[1]{\eqref{eq:#1}}

\begin{document}

\title{DRIT++: Diverse Image-to-Image Translation via Disentangled Representations}

\author{Hsin-Ying Lee*\thanks{* Equal contribution} \and
        Hung-Yu Tseng* \and
        Qi Mao*        \and
        Jia-Bin Huang \and 
        Yu-Ding Lu \and 
        Maneesh Singh \and
        Ming-Hsuan Yang
}


\institute{Hsin-Ying Lee, Hung-Yu Tseng, Yu-Ding Lu, and Ming-Hsuan Yang \at
               Electrical Engineering and Computer Science, University of California at Merced, Merced, CA 95343 \\
              %Tel.: +209-201-5123\\
              \email{\{hlee246, htseng6, ylu52, mhyang\}@ucmerced.edu}           %  \\
           \and
           Qi Mao \at
              Electrical Engineering and Computer Science, Peking University, Beijing, China\\
              \email{qimao@pku.edu.cn}
           \and
           Jia-Bin Huang \at
             Electrical and Computer Engineering, Virginia Tech, Blacksburg, VA 24060\\
             \email{jbhuang@vt.edu}
          \and
          Maneesh Singh \at
          Verisk Analytics, Jersey City, NJ 07310\\
          \email{maneesh.singh@verisk.com}
}

\date{Received: date / Accepted: date}

\maketitle

\begin{figure*}[h]
\centering 
\includegraphics[width=\linewidth]{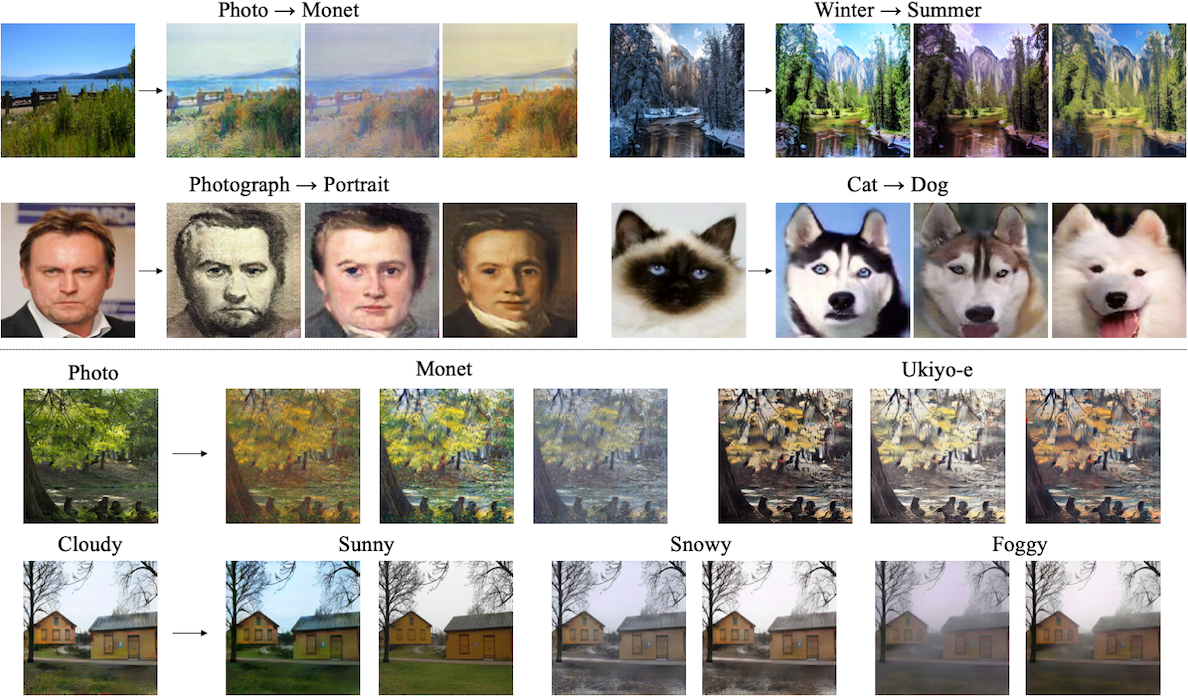}
    \caption{\textbf{Unpaired diverse image-to-image translation.}
    (\textit{\(Top\)}) Our model learns to perform diverse translation between two collections of images without aligned training pairs.
    (\textit{\(Bottom\)}) Multi-domain image-to-image translation.
    }
    %%\vspace{-13mm}
    \label{figure:teaser}
\end{figure*}

\begin{abstract}
Image-to-image translation aims to learn the mapping between two visual domains.
There are two main challenges for this task: 
1) lack of aligned training pairs and 
2) multiple possible outputs from a single input image.
In this work, we present an approach based on disentangled representation for generating diverse outputs without paired training images.
To synthesize diverse outputs, we propose to embed images onto two spaces: a domain-invariant content space capturing shared information across domains and a domain-specific attribute space.
Our model takes the encoded content features extracted from a given input and attribute vectors sampled from the attribute space to synthesize diverse outputs at test time.
To handle unpaired training data, we introduce a cross-cycle consistency loss based on disentangled representations.
Qualitative results show that our model can generate diverse and realistic images on a wide range of tasks without paired training data.
For quantitative evaluations, we measure realism with user study and Fr\'{e}chet inception distance, and measure diversity with the perceptual distance metric, Jensen-Shannon divergence, and number of statistically-different bins.
\end{abstract}

%%%%%%%%%%%%%%%%%%%%%%%%%%%%%%%%%%%%%%%%%%%%%%%%%%%%%%%%%%
%%%%%%%%%%%%%%%%%%    Introduction    %%%%%%%%%%%%%%%%%%%%
%%%%%%%%%%%%%%%%%%%%%%%%%%%%%%%%%%%%%%%%%%%%%%%%%%%%%%%%%%
%-------------------------------------------------------------
\begin{figure*}[t]
\centering
	\begin{minipage}{0.31\textwidth}
    \subfloat[CycleGAN~\cite{zhu2017cyclegan}]{
    \includegraphics[width=\linewidth]{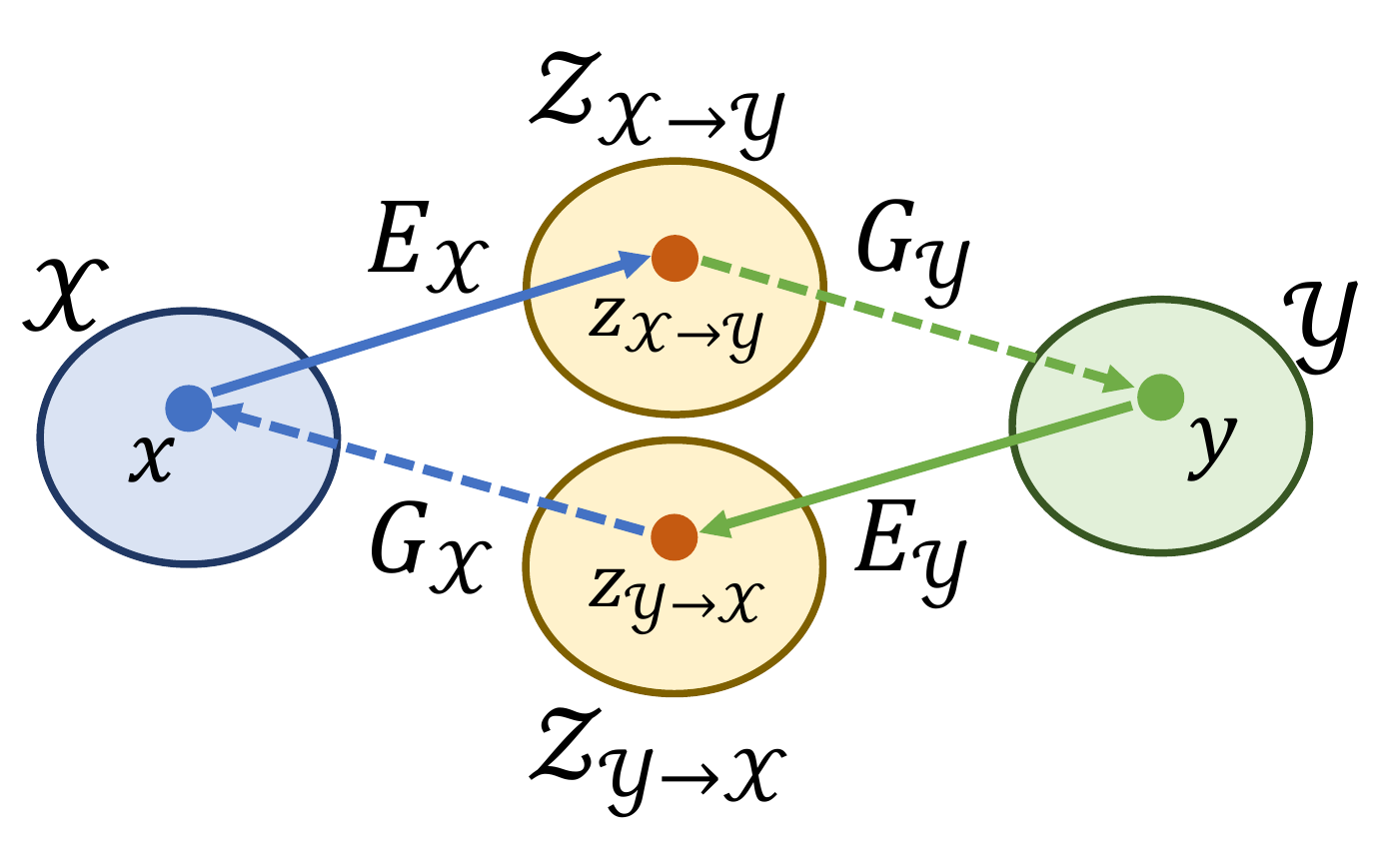}
    }
    \end{minipage}
	\hfill
    \begin{minipage}{0.31\textwidth}
	\subfloat[UNIT~\cite{liu2017unit}]{
    \includegraphics[width=\linewidth]{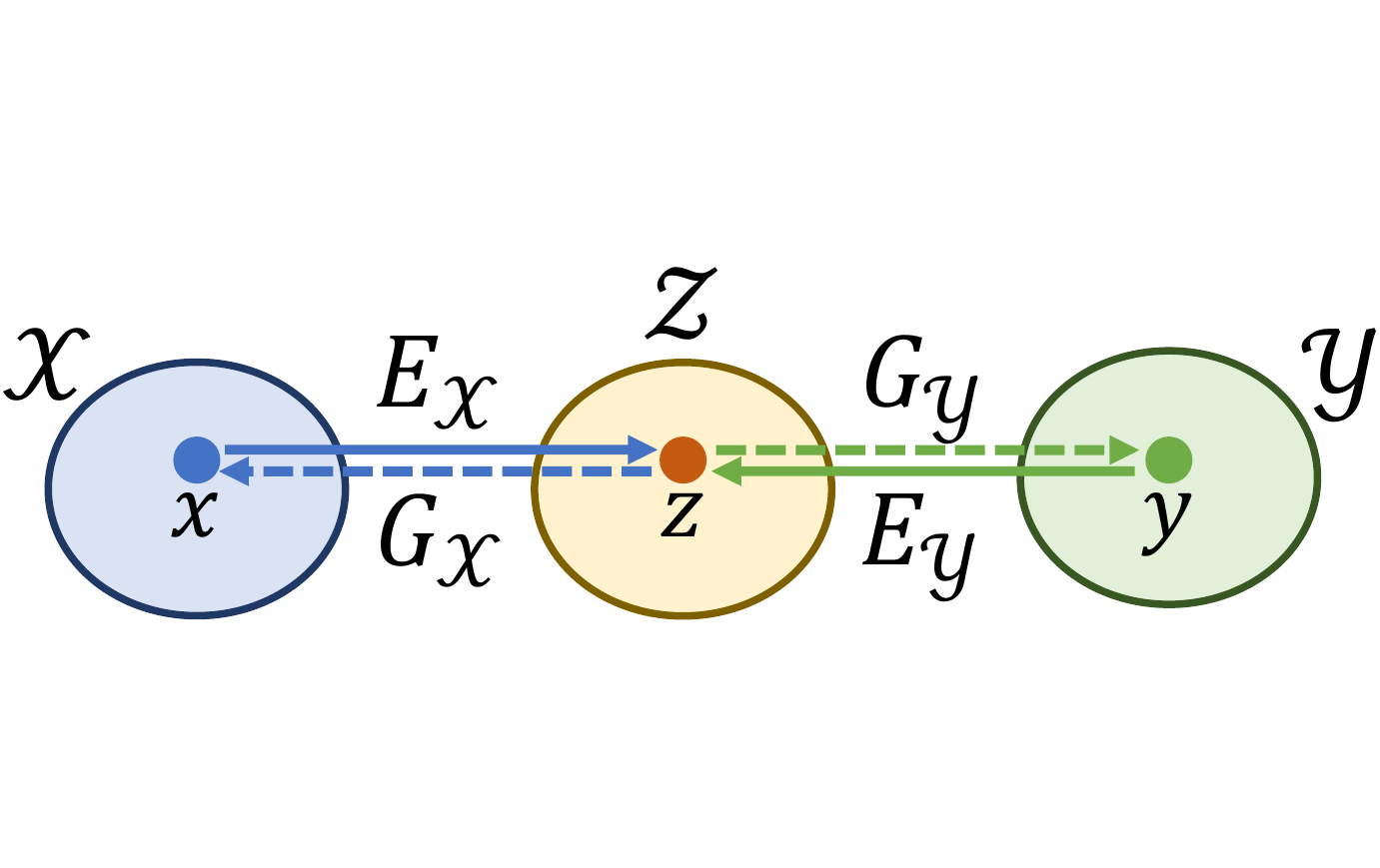}
    }
    \end{minipage}
	\hfill	
	\begin{minipage}{0.31\textwidth}
    \subfloat[MUNIT~\cite{huang2018munit}, DRIT~\cite{DRIT}]{
    \includegraphics[width=\linewidth]{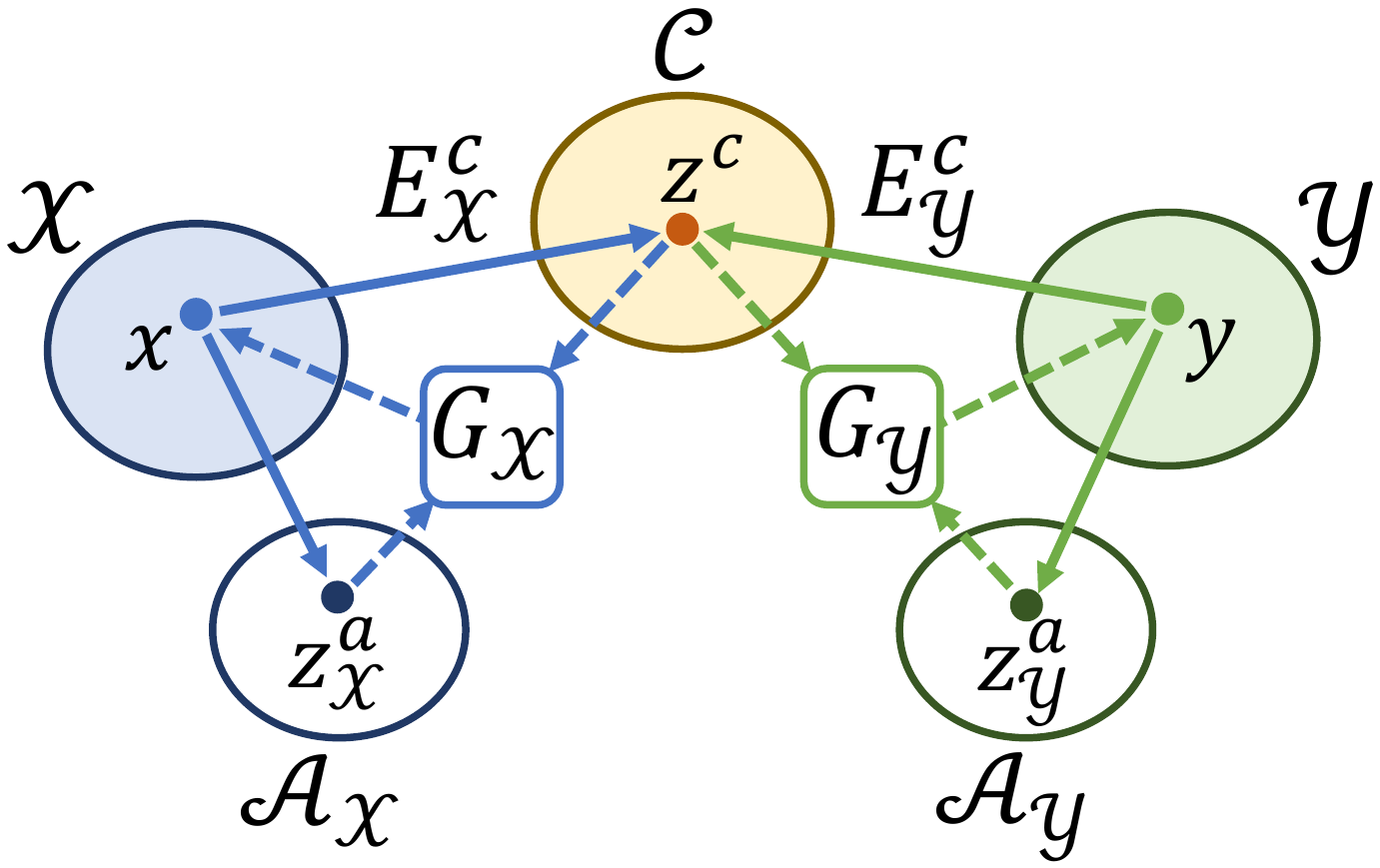}
    }
    \end{minipage}
    \caption{ 
    \textbf{Comparisons of unsupervised I2I translation methods.}
    Denote $x$ and $y$ as images in domain $\mathcal{X}$ and $\mathcal{Y}$: 
    (a) CycleGAN~\cite{zhu2017cyclegan} maps $x$ and $y$ onto \emph{separated} latent spaces.
    (b) UNIT~\cite{liu2017unit} assumes $x$ and $y$ can be mapped onto a  \emph{shared} latent space.
    (c) Our approach disentangles the latent spaces of $x$ and $y$ into a shared content space $\mathcal{C}$ and an attribute space $\mathcal{A}$ of each domain.
    }
    %\vspace{\figmargin}
    \label{figure:assumption}
\end{figure*}

%-------------------------------------------------------------

% \vspace{\secmargin}
\section{Introduction}
\label{sec:introduction}
% \vspace{\secmargin}
Image-to-Image (I2I) translation aims to learn the mapping between different visual domains. 
Numerous vision and graphics problems can be formulated as I2I translation problems, such as colorization~\cite{larsson2016colorization,zhang2016colorization2} (grayscale $\rightarrow$ color), super-resolution~\cite{lai2017deep,ledig2016photo,li2016deep,li2019joint} (low-resolution $\rightarrow$ high-resolution), and photorealistic image synthesis~\cite{chen2017photographic,park2019semantic,wang2017pix2pixhd} (label $\rightarrow$ image).
In addition, I2I translation can be applied to synthesize images for domain adaptation~\cite{bousmalis2017unsupervisedda,Chen2019CrDoCo,hoffman2017cycada,murez2018image,shrivastava2017apple}.

%\Paragraph{I2I translation is difficult for two main reasons. First, paired data is usually not applicable and the translation is even not well-defined. Second, many mappings are by nature one-to-many, which is hard to model especially on high dimensional data.}
Learning the mapping between two visual domains is challenging for two main reasons. 
First, aligned training image pairs are either difficult to collect (\eg day scene $\leftrightarrow$ night scene) or do not exist (\eg artwork $\leftrightarrow$ real photo).
Second, many such mappings are inherently multimodal --- a single input may correspond to multiple possible outputs.
To handle multimodal translation, one possible approach is to inject a random noise vector to the generator for modeling the multimodal data distribution in the target domain.
However, mode collapse may still occur easily since the generator often ignores the additional noise vectors.

%\Paragraph{Pix2pix can learn the mapping but generate single output conditioned on input and need paired data. BicycleGAN can generate multimodal results but still need paired data. On the other hand, CycleGan and UNIT can learn from unpaired data but fail to have diverse outputs.}
Several recent efforts have been made to address these issues.
The {Pix2pix}~\cite{isola2017pix2pix} method applies conditional generative adversarial network to I2I translation problems.
Nevertheless, the training process requires paired data.
A number of recent approaches~\cite{choi2017stargan,liu2017unit,taigman2016unsupervised,yi2017dualgan,zhu2017cyclegan} relax the dependency on paired training data for learning I2I translation. 
These methods, however, generate a single output conditioned on the given input image.
As shown in~\cite{isola2017pix2pix,zhu2017bicyclegan}, the strategy of incorporating noise vectors as additional inputs to the generator does not increase variations of generated outputs due to the mode collapse issue.
The generators in these methods are likely to overlook the added noise vectors.
Most recently, the {BicycleGAN}~\cite{zhu2017bicyclegan} algorithm tackles the problem of generating diverse outputs in I2I translation by encouraging the one-to-one relationship between the output and the latent vector.
Nevertheless, the training process of {BicycleGAN} requires paired images.

%\Paragraph{ High-level intuition: We propose Disentangled Representation Image-to-Image Translation that disentangles two spaces }
In this paper, we propose a disentangled representation framework for learning to generate \emph{diverse} outputs with \emph{unpaired} training data.
We propose to embed images onto two spaces: 
1) a domain-invariant content space and 2) a domain-specific attribute space as shown in \figref{assumption}.
Our generator learns to perform I2I translation conditioned on content features and a latent attribute vector. 
The domain-specific attribute space aims to model variations within a domain given the same content, while the domain-invariant content space captures information across domains.
We disentangle the representations by applying a content adversarial loss to encourage the content features \emph{not} to carry domain-specific cues, and a latent regression loss to encourage the invertible mapping between the latent attribute vectors and the corresponding outputs.
To handle unpaired datasets, we propose a \textit{cross-cycle consistency loss} using the proposed disentangled representations.
Given a pair of unaligned images, we first perform a cross-domain mapping to obtain intermediate results by swapping the attribute vectors from both images.
We can then reconstruct the original input image pair by applying the cross-domain mapping one more time and use the proposed cross-cycle consistency loss to enforce the consistency between the original and the reconstructed images. 
Furthermore, we apply the mode seeking regularization~\cite{MSGAN} to further improve the diversity of generated images.
At test time, we can use either 
1) randomly sampled vectors from the attribute space to generate diverse outputs or 
2) the transferred attribute vectors extracted from existing images for example-guided translation.
\figref{teaser} shows examples of diverse outputs produced by our model.
% of the two testing modes.
% \jiabin{
% Inconsistent. Figure 1 no longer show the two testing modes.
% }

% ===\
%\Paragraph{Qualitative. Quantitative (diversity, realism). Domain adaptation}
%MH: evaluate evaluation?
%We evaluate the proposed model through extensive qualitative and quantitative evaluation.
We evaluate the proposed model with extensive qualitative and quantitative experiments. 
For various I2I tasks, we show diverse translation results with randomly sampled attribute vectors and example-guided translation with transferred attribute vectors from existing images.
In addition to the common dual-domain image-to-image translation, we extend our proposed framework to the more general multi-domain image-to-image translation and demonstrate diverse translation among domains.
%
%Mode seeking regularization encourages generators to explore different modes during training by maximizing the ratio of the distance between generated images with respect to the corresponding latent codes.
%
We measure realism of our results with a user study and the Fr\'{e}chet inception distance (FID)~\cite{heusel2017gans}, and evaluate diversity using perceptual distance metrics~\cite{zhang2018perceptual}.
However, the diversity metric alone does not effectively measure similarity between the distribution of generated images and the distribution of real data.
Therefore, we use the Jensen-Shannon Divergence (JSD) distance which measures the similarity between distributions, and the Number of Statistically-Different Bins (NDB)~\cite{richardson2018NDB} metric which determines the relative proportions of samples within clusters predetermined by real data.
%Finally, we demonstrate the potential application of unsupervised domain adaptation.
%
%On the tasks of adapting domains from MNIST~\cite{lecun1998MNIST} to MNIST-M~\cite{ganin2016MNISTM} and Synthetic Cropped LineMod to Cropped LineMod~\cite{hinterstoisser2012linemod,wohlhart2015croplinemod}, we show competitive performance against state-of-the-art domain adaptation methods.

%-------------------------------------------------------------
%\begin{table*}[t]
%	\caption{\textbf{Feature-by-feature comparison of image-to-image translation networks.} Our model achieves multimodal translation without using aligned training image pairs.}
 %	\label{tab:related_work}
%	\centering
%	\begin{tabular}{l ccccc} 
%    	\toprule
%Method & 
%Pix2Pix~\cite{isola2017pix2pix} \ \ & CycleGAN~\cite{zhu2017cyclegan} \ \ & 
%UNIT~\cite{liu2017unit} \ \ &
%BicycleGAN~\cite{zhu2017bicyclegan} \ \ & 
%Ours \ \  \\
%        \midrule
%        Unpaired& - & \cmark& \cmark&-& \cmark\\
%        Multimodal&-&-&-& \cmark& \cmark\\
%		\bottomrule
%	\end{tabular}
%    \vspace{\tabmargin}
%\end{table*}
%-------------------------------------------------------------

%\Paragraph{We make the following contributions in this work: 1. First to achieve unpaired multimodal image-to-image translation by disentangling content and style. 2. We propose a cross-cycle consistency loss and content adversarial loss 3. Can be used in domain adaptation.}
We make the following contributions in this work:

1) We introduce a disentangled representation framework for image-to-image translation. 
We apply a content discriminator to facilitate the factorization of domain-invariant content space and domain-specific attribute space, and a cross-cycle consistency loss that allows us to train the model with unpaired data.

2)
Extensive qualitative and quantitative experiments show that our model performs favorably against existing I2I models.
Images generated by our model are both diverse and realistic.

3)
The proposed disentangled representation and  cross-cycle consistency can be applied to multi-domain image-to-image translation for generating diverse images.
%We demonstrate the application of our model on unsupervised domain adaptation.
%
%We achieve competitive results on both the MNIST-M and the Cropped LineMod datasets.
%

%MH: This paragraph usually goes to cover letter (especially for IJCV).
% ===\
%\Paragraph{Original DRIT}

%A preliminary version of this work is presented in~\cite{DRIT}.
%
%In this work, we make significant extensions and summarize the differences as follows.
%
%First, we incorporate the recently proposed mode seeking regularization~\cite{MSGAN} to further improve the diversity.
%
%Experimental results show that the mode seeking regularization improves diversity without the loss of visual quality.
%
%Second, we apply the proposed framework to the multi-domain image-to-image translation task.
%
%Existing work on multi-domain image-to-image translation~\cite{StarGAN2018,UFDN2018} only perform one-to-one mapping among domains.
%
%We leverage the disentangle representation and the cross-cycle consistency to achieve multimodal multi-domain image-to-image translation.
%
%Third, we demonstrate the capability of generating high-resolution diverse image-to-image translation.
%
%We adopt a multi-scale generator-discriminator structure to stabilize the training and enhance the quality of the synthesized images.
%
%The experiments on the street scene dataset validate the applicability of the proposed framework on high-resolution translation task.

%%%%%%%%%%%%%%%%%%%%%%%%%%%%%%%%%%%%%%%%%%%%%%%%%%%%%%%%%%
%%%%%%%%%%%%%%%%%%    Related Work    %%%%%%%%%%%%%%%%%%%%
%%%%%%%%%%%%%%%%%%%%%%%%%%%%%%%%%%%%%%%%%%%%%%%%%%%%%%%%%%

\section{Related Work}
\label{sec:related}
%%\vspace{-1mm}
% \vspace{\secmargin}
%%%%%%%%%%%%%%%%%%%%
%%%Generative Adversarial Network
%%%%%%%%%%%%%%%%%%%%
\Paragraph{Generative adversarial networks.}
The recent years have witnessed rapid advances of generative adversarial networks (GANs)~\cite{arjovsky2017wgan,goodfellow2014GAN,radford2016dcgan} for image generation.
The core idea of GANs lies in the adversarial loss that enforces the distribution of generated images to match that of the target domain.
The generators in GANs can map from noise vectors to realistic images.
Several recent efforts exploit \emph{conditional} GAN in various contexts including conditioned on text~\cite{reed2016text2img}, audio~\cite{lee2019dancing2music}, low-resolution images~\cite{ledig2016photo}, human pose~\cite{ma2017pose,albahar2019guided}, video frames~\cite{vondrick2016videogan}, and image~\cite{isola2017pix2pix}.
Our work focuses on using GAN conditioned on an input image.
In contrast to several existing conditional GAN frameworks that require paired training data, our model generates diverse outputs without paired data.
As such, our method has wider applicability to problems where paired training datasets are scarce or not available.

\vspace{\paramargin}
%%%%%%%%%%%%%%%%%%%%
%%% Image-to-Image Translation
%%%%%%%%%%%%%%%%%%%%
\Paragraph{Image-to-image translation.}
I2I translation aims to learn the mapping from a source image domain to a target image domain.
The Pix2pix~\cite{isola2017pix2pix} method applies a conditional GAN to model the mapping function.
Although high-quality results have been shown, the model training requires paired training data. 
To train with unpaired data, the CycleGAN~\cite{zhu2017cyclegan}, DiscoGAN~\cite{kim2017discogan}, and UNIT~\cite{liu2017unit} schemes leverage cycle consistency to regularize the training.
However, these methods perform generation conditioned solely on an input image and thus produce one single output.
Simply injecting a noise vector to a generator is usually not an effective solution to achieve multimodal generation due to the lack of regularization between the noise vectors and the target domain. 
On the other hand, the BicycleGAN~\cite{zhu2017bicyclegan} algorithm enforces the bijection mapping between the latent and target space to tackle the mode collapse problem.
Nevertheless, the method is only applicable to problems with paired training data. 
%
% Table~\ref{tab:related_work} shows a feature-by-feature comparison among various I2I models. 
%
Unlike existing work, our method enables I2I translation with diverse outputs in the absence of paired training data.

We note several concurrent methods~\cite{almahairi2018augmented,cao2018dida,huang2018munit,lin2019explore,lin2018conditional,ma2018exemplar} (all independently developed) also adopt  disentangled representations similar to our work for learning diverse I2I translation from unpaired training data.
%
%We encourage the readers to review these works for a complete picture.
%
Furthermore, several  approaches~\cite{StarGAN2018,UFDN2018} extend the conventional dual-domain I2I to general multi-domain settings.
However, these methods can only achieve one-to-one mapping among domains.

%-------------------------------------------------------------
\begin{figure*}[t]
	\centering
	\subfloat[Training with unpaired images]{%
		\includegraphics[width=0.95\linewidth]{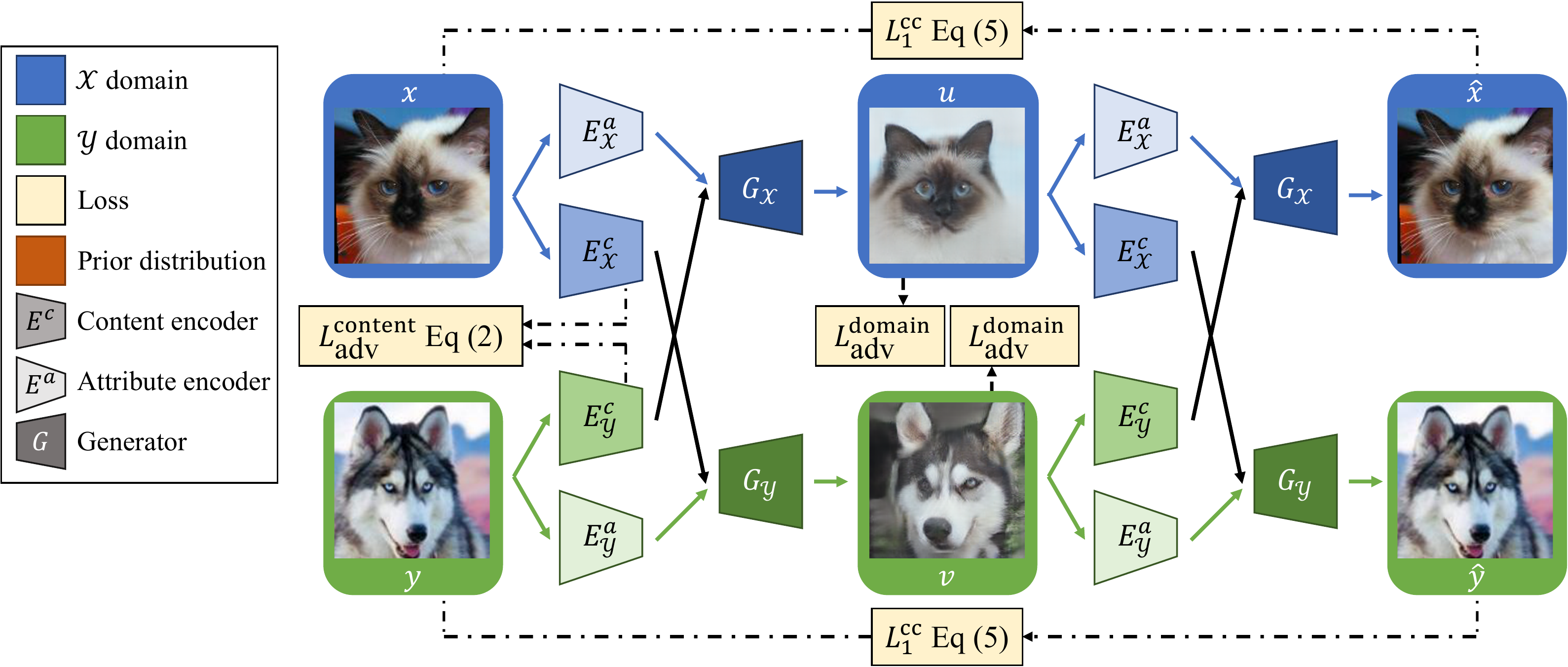}%
	}
    
    \begin{minipage}[b]{0.491\textwidth}
    \subfloat[Testing with random attributes]{%
		\includegraphics[width=\linewidth]{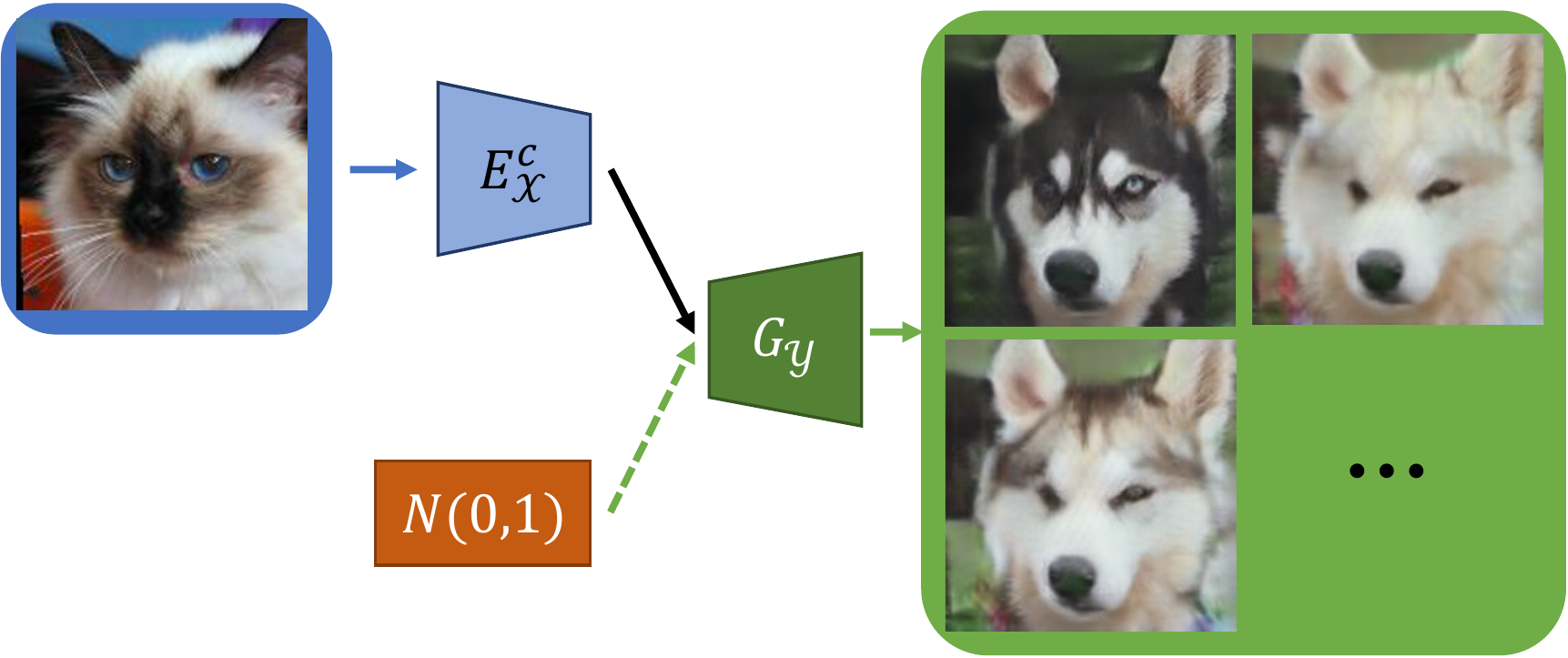}%
	}
    \end{minipage}
    \hspace{4mm}
    \begin{minipage}[b]{0.409\textwidth}
    \subfloat[Testing with a given attribute ]{%
		\includegraphics[width=\linewidth]{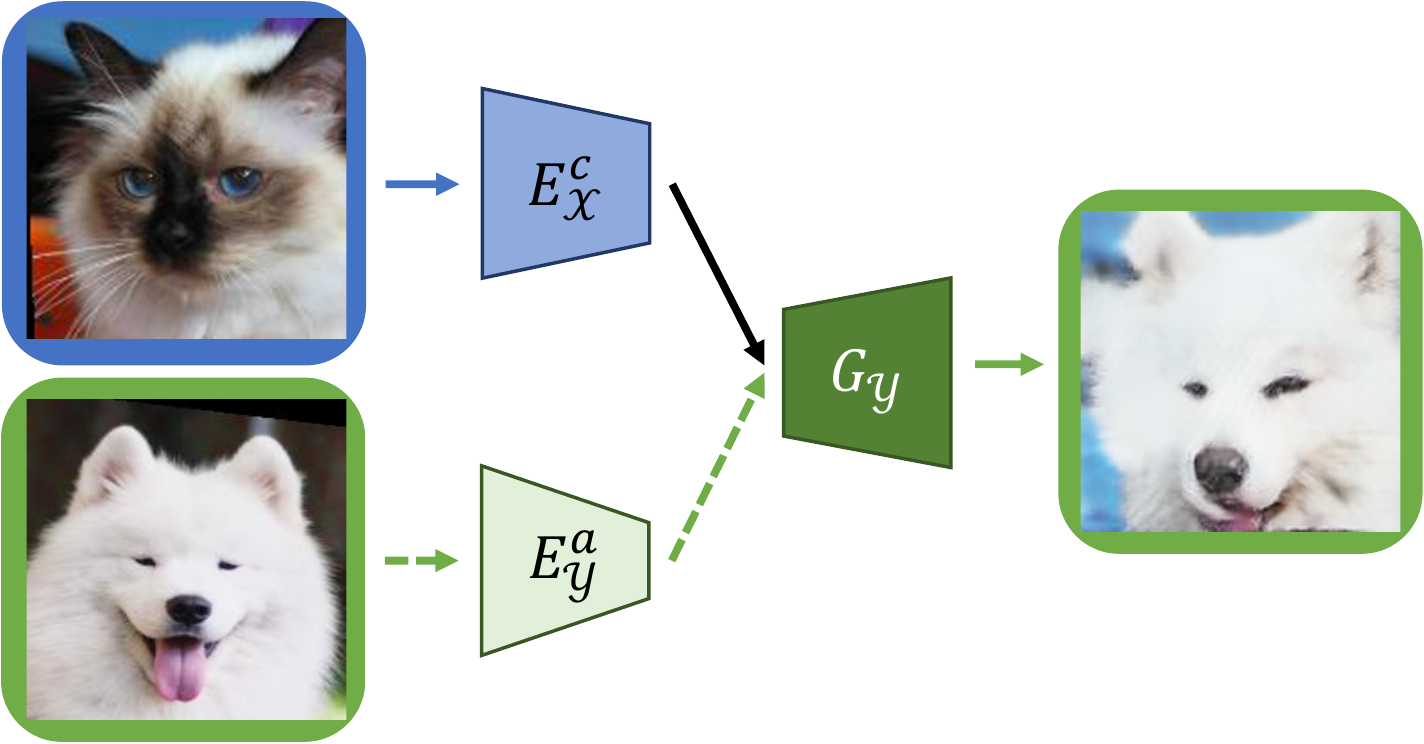}%
	}
    \end{minipage}
    \caption{\textbf{Method overview.} (a) With the proposed content adversarial loss $L_\mathrm{adv}^\mathrm{content}$ (Section~\ref{subsec:conadv}) and the cross-cycle consistency loss $L_1^\mathrm{cc}$ (Section~\ref{subsec:crosscycle}), we are able to learn the multimodal mapping between the domain $\mathcal{X}$ and $\mathcal{Y}$ with unpaired data. 
    Thanks to the proposed disentangled representation, we can generate output images conditioned on either (b) random attributes or (c) a given attribute at test time.}
    \label{figure:architecture}
    %\vspace{\figmargin}
\end{figure*}
%-------------------------------------------------------------

\vspace{\paramargin}
%%%%%%%%%%%%%%%%%%%%
%%% Disentangled Representations
%%%%%%%%%%%%%%%%%%%%
\Paragraph{Disentangled representations.}
The task of learning disentangled representation aims at modeling the factors of data variations.
Previous work makes use of labeled data to factorize representations into class-related and class-independent components~\cite{cheung2014discovering,kingma2014semi,makhzani2015adversarial,mathieu2016disentangling}.
Recently, numerous unsupervised methods have been developed~\cite{chen2016infogan,denton2017unsupervised} to learn disentangled representations. 
The InfoGAN~\cite{chen2016infogan} algorithm achieves disentanglement by maximizing the mutual information between latent variables and data variation.
Similar to DrNet~\cite{denton2017unsupervised} that separates time-independent and time-varying components with an adversarial loss, we apply a content adversarial loss to disentangle an image into domain-invariant and domain-specific representations to facilitate learning diverse cross-domain mappings. 

\vspace{\paramargin}
%%%%%%%%%%%%%%%%%%%%
%%% Domain Adaptation
%%%%%%%%%%%%%%%%%%%%
%\Paragraph{Domain adaptation.}
%Domain adaptation techniques focus on addressing the domain-shift problem between a source and a target domain.
%
%Domain Adversarial Neural Network (DANN)~\cite{ganin2015unsupervised,ganin2016domain} and its variants~\cite{tzeng2014deep,bousmalis2016domain,Tsai_adaptseg_2018} tackle domain adaptation through learning domain-invariant features.
%
%Sun~\etal~\cite{sun2016return} aims to map features in the source domain to those in the target domain. 
%
%I2I translation has been recently applied to produce simulated images in the target domain by translating images from the source domain~\cite{ganin2015unsupervised,hoffman2017cycada}.
%
%Different from the aforementioned I2I based domain adaptation algorithms, our method does not utilize source domain annotations for I2I translation. 

%%%%%%%%%%%%%%%%%%%%%%%%%%%%%%%%%%%%%%%%%%%%%%%%%%%%%%%%%%
%%%%%%%%%%%%%%%%%%%%   Framework    %%%%%%%%%%%%%%%%%%%%%%
%%%%%%%%%%%%%%%%%%%%%%%%%%%%%%%%%%%%%%%%%%%%%%%%%%%%%%%%%%

\section{Disentangled Representation for I2I Translation}
\label{sec:framework}
% \vspace{\secmargin}
%
Our goal is to learn a multimodal mapping between two visual domains $\mathcal{X} \subset \mathbb{R}^{H\times W \times 3}$ and $\mathcal{Y} \subset \mathbb{R}^{H\times W \times 3}$ without paired training data.
As illustrated in~\figref{architecture}, our framework consists of content encoders $\{E^c_\mathcal{X}, E^c_\mathcal{Y}\}$, attribute encoders $\{E^a_\mathcal{X}, E^a_\mathcal{Y}\}$, generators $\{G_\mathcal{X}, G_\mathcal{Y}\}$, and domain discriminators $\{D_\mathcal{X}, D_\mathcal{Y}\}$ for both domains, and a content discriminators $D_\mathrm{adv}^\mathrm{c}$.
Taking domain $\mathcal{X}$ as an example, the content encoder $E^c_\mathcal{X}$ maps images onto a shared, domain-invariant content space ($E^c_\mathcal{X}:\mathcal{X}\to \mathcal{C}$) and the attribute encoder $E^a_\mathcal{X}$ maps images onto a domain-specific attribute space ($E^a_\mathcal{X}:\mathcal{X}\to \mathcal{A}_\mathcal{X}$).
The generator $G_\mathcal{X}$ synthesizes images conditioned on both content and attribute vectors ($G_\mathcal{X}:\{\mathcal{C}, \mathcal{A}_\mathcal{X}\} \to \mathcal{X} $).
The discriminator $D_\mathcal{X}$ aims to discriminate between real images and translated images in the domain $\mathcal{X}$.
In addition, the content discriminator $D^c$ is trained to distinguish the extracted content representations between two domains.
To synthesize multimodal outputs at test time, we regularize the attribute vectors so that they can be drawn from a prior Gaussian distribution $\mathnormal{N}(0,1)$. 

%In this section, we first discuss the strategies used to disentangle the content and attribute representations in \subsecref{conadv} and then introduce the proposed cross-cycle consistency loss that enables the training on unpaired data in \subsecref{crosscycle}. 
%
%Finally, we detail the loss functions in \subsecref{learn}.

%-------------------------------------------------------------
\begin{figure*}[t]
	\centering
		\includegraphics[width=\linewidth]{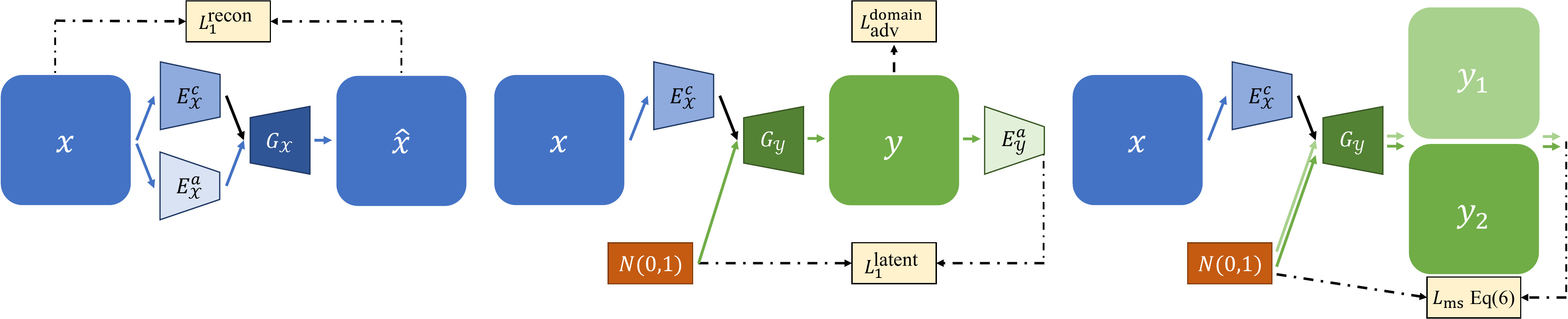}%
	\caption{\textbf{Additional loss functions.} 
    In addition to the cross-cycle reconstruction loss $L_1^{\mathrm{cc}}$ and the content adversarial loss $L_{\mathrm{adv}}^\mathrm{content}$ described in Figure~\ref{figure:architecture}, we apply several additional loss functions in our training process. 
    The self-reconstruction loss $L_1^{\mathrm{recon}}$ facilitates training with self-reconstruction;
    the KL loss $L_{\mathrm{KL}}$ aims to align the attribute representation with a prior Gaussian distribution;
    the adversarial loss $L_{\mathrm{adv}}^{\mathrm{domain}}$ encourages $G$ to generate realistic images in each domain; 
    and the latent regression loss $L_1^{\mathrm{latent}}$ enforces the reconstruction on the latent attribute vector.
    Finally, the mode seeking regularization $L_{\mathrm{ms}}$ further improves the diversity.
    More details can be found in Section~\ref{subsec:learn} ans Section~\ref{subsec:msgan}.}
	\label{figure:loss}
    %\vspace{\figmargin}
\end{figure*}
%-------------------------------------------------------------

\subsection{Disentangle Content and Attribute Representations}
\label{subsec:conadv}
% \vspace{\subsecmargin}
%
Our approach embeds input images onto a shared content space $\mathcal{C}$, and domain-specific attribute spaces, $\mathcal{A}_\mathcal{X}$ and $\mathcal{A}_\mathcal{Y}$.
Intuitively, the content encoders should encode the common information that is \emph{shared} between domains onto $\mathcal{C}$, while the attribute encoders should map the remaining domain-specific information onto $\mathcal{A}_\mathcal{X}$ and $\mathcal{A}_\mathcal{Y}$.
\vspace{\eqmargin}
\begin{equation}
\begin{aligned}
&\{z_x^{c},z_x^{a}\} = \{{E^c_\mathcal{X}}(x), {E^a_\mathcal{X}}(x)\}\quad&& z_x^{c}\in \mathcal{C}, z_x^{a}\in \mathcal{A}_\mathcal{X},\\
&\{z_y^{c},z_y^{a}\} = \{{E^c_\mathcal{Y}}(y), {E^a_\mathcal{Y}}(y)\}\quad&& z_y^{c}\in \mathcal{C}, z_y^{a}\in \mathcal{A}_\mathcal{Y}.
\end{aligned}
\end{equation}
\vspace{\eqmargin}

To achieve representation disentanglement, we apply two strategies: weight-sharing and a content discriminator.
First, similar to~\cite{liu2017unit}, based on the assumption that two domains share a common latent space, we share the weight between the last layer of $E^c_\mathcal{X}$ and $E^c_\mathcal{Y}$ and the first layer of $G_\mathcal{X}$ and $G_\mathcal{Y}$.
Through weight sharing, we enforce the content representation to be mapped onto the same space.
However, sharing the same high-level mapping functions does not guarantee the same content representations encode the same information for both domains.
Thus, we propose a content discriminator $D^c$ which aims to distinguish the domain membership of the encoded content features $z_x^{c}$ and $z_y^{c}$.
On the other hand, content encoders learn to produce encoded content representations whose domain membership cannot be distinguished by the content discriminator $D^c$.
We express this content adversarial loss as:
\vspace{\eqmargin}
\begin{equation}
\begin{aligned}
L_{\mathrm{adv}}^{\mathrm{content}}&(E^c_\mathcal{X},E^c_\mathcal{Y}, D^c) =\\ &\mathbb{E}_{x}[\frac{1}{2}\log{D^c(E^c_\mathcal{X}(x))}+\frac{1}{2}\log{(1-D^c(E^c_\mathcal{X}(x)))]}\\  +& \mathbb{E}_{y}[\frac{1}{2}\log{D^c(E^c_\mathcal{Y}(y))}+\frac{1}{2}\log{(1-D^c(E^c_\mathcal{Y}(y)))}].
\end{aligned}
\end{equation}
\vspace{\eqmargin}
\vspace{\eqmargin}

% \vspace{\subsecmargin}
\subsection{Cross-cycle Consistency Loss}
\label{subsec:crosscycle}
% \vspace{\subsecmargin}
%
With the disentangled representation where the content space is shared among domains and the attribute space encodes intra-domain variations, we can perform I2I translation by combining a content representation from an arbitrary image and an attribute representation from an image of the target domain.
We leverage this property and propose a \textit{cross-cycle consistency}.
In contrast to cycle consistency constraint in~\cite{zhu2017cyclegan} (\ie{$\mathcal{X} \to \mathcal{Y} \to \mathcal{X}$}) which assumes one-to-one mapping between the two domains, the proposed cross-cycle constraint exploit the disentangled content and attribute representations for cyclic reconstruction.

Our cross-cycle constraint consists of two stages of I2I translation.

\vspace{\paramargin}
\Paragraph{Forward translation.} Given a non-corresponding pair of images $x$ and $y$,  we encode them into $\{z_x^{c}, z_x^{a}\}$ and $\{z_y^{c}, z_y^{a}\}$.
We then perform the first translation by swapping the attribute representation (\ie $z_x^{a}$ and $z_y^{a}$) to generate $\{u,v\}$, where $u\in \mathcal{X}, v \in \mathcal{Y}$.
\vspace{\eqmargin}
\begin{equation}
\begin{aligned}
u = G_\mathcal{X}(z_y^c, z_x^a)\quad
v = G_\mathcal{Y}(z_x^c, z_y^a).
\end{aligned}
\end{equation}
\vspace{\eqmargin}

\vspace{\paramargin}
\Paragraph{Backward translation.} After encoding $u$ and $v$ into $\{z_u^c,z_u^a\}$ and $\{z_v^c,z_v^a\}$, we perform the second translation by once again swapping the attribute representation (\ie $z_u^a$ and $z_v^a$).
\vspace{\eqmargin}
\begin{equation}
\begin{aligned}[c]
\hat{x} = G_\mathcal{X}(z_v^c, z_u^a)\quad
\hat{y} = G_\mathcal{Y}(z_u^c, z_v^a).
\end{aligned}
\end{equation}
\vspace{\eqmargin}

Here, after two I2I translation stages, the translation should reconstruct the original images $x$ and $y$ (as illustrated in \figref{architecture}).
To enforce this constraint, we formulate the \textit{cross-cycle consistency loss} as:

\vspace{\eqmargin}
\begin{equation}
\begin{aligned}
L_1^{\mathrm{cc}}(G_\mathcal{X},G_\mathcal{Y},&E_\mathcal{X}^c,E_\mathcal{Y}^c,E_\mathcal{X}^a,E_\mathcal{Y}^a) =\\ 
\mathbb{E}_{x,y}[&\lVert G_\mathcal{X}(E_\mathcal{Y}^c(v),E_\mathcal{X}^a(u) )-x \lVert_{1} \\
+ &\lVert G_\mathcal{Y}(E_\mathcal{X}^c(u),E_\mathcal{Y}^a(v) )-y \lVert_{1}],\\
\end{aligned}
\end{equation}
%\vspace{\eqmargin}
where $u=G_\mathcal{X}(E_\mathcal{Y}^c(y),E_\mathcal{X}^a(x))$ and $v=G_\mathcal{Y}(E_\mathcal{X}^c(x),E_\mathcal{Y}^a(y))$, respectively. 

\subsection{Other Loss Functions}
\label{subsec:learn}
% \vspace{\subsecmargin}
In addition to the proposed content adversarial loss and cross-cycle consistency loss, we also use several other loss functions to facilitate network training.
We illustrate these additional losses in \figref{loss}. 
Starting from the top-right, in the counter-clockwise order:

\vspace{\paramargin}
\Paragraph{Domain adversarial loss.}
We impose an adversarial loss $L_{\mathrm{adv}}^{\mathrm{domain}}$ where $D_\mathcal{X}$ and $D_\mathcal{Y}$ attempt to discriminate between real images and generated images in each domain, while $G_\mathcal{X}$ and $G_\mathcal{Y}$ attempt to generate realistic images.

\vspace{\paramargin}
\Paragraph{Self-reconstruction loss.}
In addition to the cross-cycle reconstruction, we apply a self-reconstruction loss $L_1^{\mathrm{rec}}$ to facilitate the training process. 
With encoded content and attribute features $\{z_x^c, z_x^a\}$ and $\{z_y^c, z_y^a\}$, the decoders $G_\mathcal{X}$ and $G_\mathcal{Y}$ should decode them back to original input $x$ and $y$.
That is, $\hat{x} = G_\mathcal{X}(E_\mathcal{X}^c(x),E_\mathcal{X}^a(x) )$ and $\hat{y} = G_\mathcal{Y}(E_\mathcal{Y}^c(y),E_\mathcal{Y}^a(y) )$.
%

%\vspace{\paramargin}
%\Paragraph{KL loss.}
%In order to perform stochastic sampling at test time, we encourage the attribute representation to be as close to a prior Gaussian distribution.  
%
%We thus apply a KL loss $L_{\mathrm{KL}}= \mathbb{E}[D_{\mathrm{KL}}((z_a)\|N(0,1))]$, where $D_{\mathrm{KL}}(p\|q)=-\int{p(z)\log{\frac{p(z)}{q(z)}}\mathrm{d}z}$.

\vspace{\paramargin}
\Paragraph{Latent regression loss.}
To encourage invertible mapping between the image and the latent space, we apply a latent regression loss $L_1^{\mathrm{latent}}$ similar to~\cite{zhu2017bicyclegan}. 
We draw a latent vector $z$ from the prior Gaussian distribution as the attribute representation and attempt to reconstruct it with $\hat{z}=E_\mathcal{X}^a(G_\mathcal{X}(E_\mathcal{X}^c(x),z))$ and $\hat{z}=E_\mathcal{Y}^a(G_\mathcal{Y}(E_\mathcal{Y}^c(y),z))$.

The full objective function of our network is:
\vspace{\eqmargin}
%\begin{equation}
%\label{eq:full}
\begin{align}
&L_{D,D^c} = & \lambda_{\mathrm{adv}}^{\mathrm{content}}L_{\mathrm{adv}}^{\mathrm{c}} + \lambda_{\mathrm{adv}}^{\mathrm{domain}}L_{\mathrm{adv}}^{\mathrm{domain}},\\
&L_{G,E^c,E^a} = & -L_{D,D^c} + \lambda_1^{\mathrm{cc}}L_1^{\mathrm{cc}} + \lambda_1^{\mathrm{recon}} L_1^{\mathrm{recon}} \nonumber \\
 &  &+ \lambda_1^{\mathrm{latent}}L_1^{\mathrm{latent}},
%\min_{G,E^c,E^a}\max_{D,D^c}\quad &\lambda_{\mathrm{adv}}^{\mathrm{content}}L_{\mathrm{adv}}^{\mathrm{c}}+\lambda_1^{\mathrm{cc}}L_1^{\mathrm{cc}} + \lambda_{\mathrm{adv}}^{\mathrm{domain}}L_{\mathrm{adv}}^{\mathrm{domain}}+ \\ &\lambda_1^{\mathrm{recon}} L_1^{\mathrm{recon}}
% + \lambda_1^{\mathrm{latent}}L_1^{\mathrm{latent}}+ \lambda_{\mathrm{KL}}L_{\mathrm{KL}}
\end{align}
%\end{equation}
%\vspace{\eqmargin}
where the hyper-parameters $\lambda$s control the importance of each term. 
%

%-------------------------------------------------------------
\begin{figure*}[t]
	\centering
		\includegraphics[width=\linewidth]{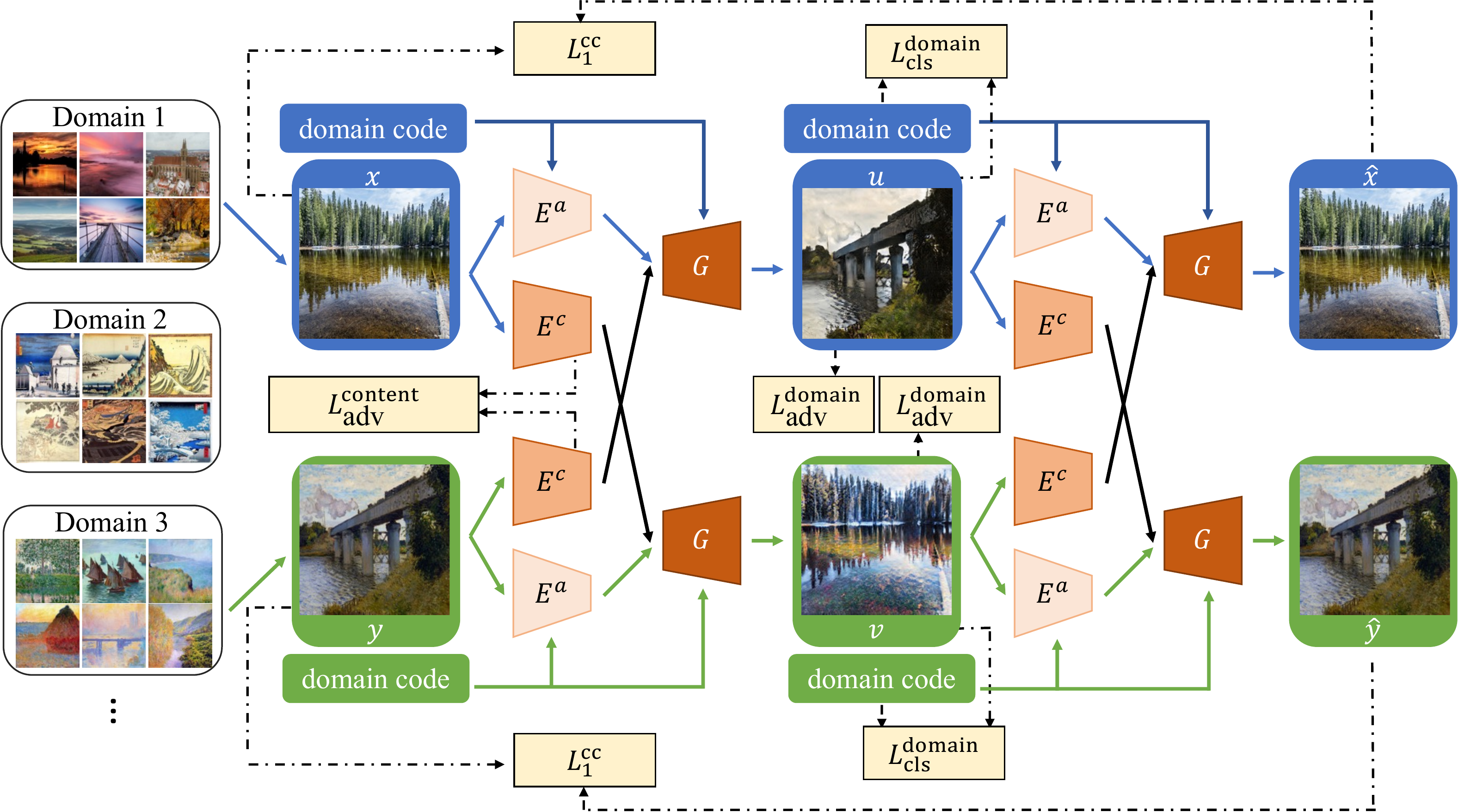}%
	\caption{\textbf{Multi-domains I2I framework.} 
	We further extend the proposed disentangle representation framework to a more general multi-domain setting.
	Different from the class-specific encoders, generators, and discriminators used in dual-domain I2I, all networks in multi-domain are shared among all domains.
	Furthermore, one-hot domain codes are used as inputs and the discriminator will perform domain classification in addition to discrimination.
	}
		\label{figure:mdmm}
    %\vspace{\figmargin}
\end{figure*}

%-------------------------------------------------------------
\subsection{Mode Seeking Regularization}
\label{subsec:msgan}
We incorporate the mode seeking regularization~\cite{MSGAN} method to alleviate the mode-collapse problem in conditional generation tasks.
% Mode seeking regularization is proposed in~\cite{MSGAN} to alleviate the mode-missing problem in conditional generation tasks.
%
Given a conditional image $\mathbf{I}$, latent vectors $\mathbf{z}_1$ and $\mathbf{z}_2$, and a conditional generator $G$, we use the mode seeking regularization term to maximize the ratio of the distance between $G(\mathbf{I}, \mathbf{z}_1)$ and $G(\mathbf{I}, \mathbf{z}_2)$ with respect to the distance between $\mathbf{z}_1$ and $\mathbf{z}_2$, 
\begin{equation}
    \mathcal{L}_{\mathrm{ms}}= \max\limits_{G} (\frac{d_\mathbf{I}(G(\mathbf{z}_1,\mathbf{I}), G(\mathbf{I}, \mathbf{z}_2))}{d_\mathbf{z}(\mathbf{z}_1,\mathbf{z}_2)}),
\label{Eq:proposed1}  
\end{equation}
where $d_{\ast}(\cdot)$ denotes the distance metric.

The regularization term can be easily incorporated into the proposed framework:
\begin{equation}
\mathcal{L}_{\mathrm{new}} = \mathcal{L}_{\mathrm{ori}} +\lambda_{\mathrm{ms}}\mathcal{L}_{\mathrm{ms}},
\end{equation}
where $\mathcal{L}_{\mathrm{ori}}$ denote the full objective.% \eqnref{full}.
%\jiabin{Where is FULL?}

%
\subsection{Multi-Domain Image-to-Image Translation}
\label{subsec:MDMM}
In addition to the translation between two domains, we apply the proposed disentangle representation to the multi-domain setting.
Different from typical I2I designed for two domains, multi-domain I2I aims to perform translation among multiple domains with a single generator $G$.

We illustrate the framework for multi-domain I2I in \figref{mdmm}.
Given $k$ domains $\{N_i\}_{i=1\sim k}$, two images ($x,y$) and their one-hot domain codes ($z^d_x,z^d_y$) are randomly sampled ($x\in N_n, y\in N_m, Z^d \subset \mathbb{R}^{k} $).
We encode the images onto a shared content space $\mathcal{C}$, and domain-specific attribute spaces $\{\mathcal{A}_i\}_{i=1\sim k}$:

\vspace{\eqmargin}
\begin{equation}
\begin{aligned}
&\{z_x^{c},z_x^{a}\} = \{{E^c}(x), E^a(x,z^d_x)\}\quad&& z_x^{c}\in \mathcal{C}, z_x^{a}\in \mathcal{A}_n,\\
&\{z_y^{c},z_y^{a}\} = \{{E^c}(y), E^a(y,z^d_y)\}\quad&& z_y^{c}\in \mathcal{C}, z_y^{a}\in \mathcal{A}_m.
\end{aligned}
\end{equation}
\vspace{\eqmargin}

We then perform the forward and backward translation similar to the dual-domain translation.
\vspace{\eqmargin}
\begin{equation}
\begin{aligned}
&u = G(z_y^c, z_x^a, z^d_x)\quad
v = G(z_x^c, z_y^a, z^d_y), \\
&\hat{x} = G(z_v^c, z_u^a, z^d_u)\quad
\hat{y} = G(z_u^c, z_v^a, z^d_v).
\end{aligned}
\end{equation}
\vspace{\eqmargin}

In addition to the loss functions used in the dual-domain translation, we leverage the discriminator $D$ as an auxiliary domain classifier.
That is, the discriminator $D$ not only aims to discriminate between real images and translated images ($D_{\mathrm{dis}}$), but also performs domain classification ($D_{\mathrm{cls}}:N_{i} \to Z^d$).
\vspace{\eqmargin}
\begin{equation}
\begin{aligned}
\mathcal{L}_{\mathrm{cls}}^{\mathrm{domain}} = &\mathbb{E}_{x,z^d_x}[-\log{D_\mathrm{cls}({z^d_x}|x)}]+ \\
&\mathbb{E}_{x,y,z^d_y}[-\log{D_\mathrm{cls}(z^d_y|G(z_x^c, z_y^a, z^d_y)}].
\end{aligned}
\end{equation}
\vspace{\eqmargin}

Thus, our new objective function is:
\vspace{\eqmargin}
\begin{equation}
%\label{eq:full}
\begin{aligned}
L_{D,D^c} = &\lambda_{\mathrm{adv}}^{\mathrm{content}}L_{\mathrm{adv}}^{\mathrm{c}} + \lambda_{\mathrm{adv}}^{\mathrm{domain}}L_{\mathrm{adv}}^{\mathrm{domain}} + \\ &\lambda_{\mathrm{cls}}^{\mathrm{domain}}\mathcal{L}_{\mathrm{cls}}^{\mathrm{domain}}, 
\end{aligned}
\end{equation}
\begin{equation}
%\label{eq:full}
\begin{aligned}
L_{G,E^c,E^a} &= -L_{D,D^c} + \lambda_1^{\mathrm{cc}}L_1^{\mathrm{cc}} + \lambda_1^{\mathrm{recon}} L_1^{\mathrm{recon}}\\
 & + \lambda_1^{\mathrm{latent}}L_1^{\mathrm{latent}}+ \lambda_{\mathrm{KL}}L_{\mathrm{KL}}+ \lambda_{\mathrm{cls}}^{\mathrm{domain}}\mathcal{L}_{\mathrm{cls}}^{\mathrm{domain}}.
%\min_{G,E^c,E^a}\max_{D,D^c}\quad &\lambda_{\mathrm{adv}}^{\mathrm{content}}L_{\mathrm{adv}}^{\mathrm{c}}+\lambda_1^{\mathrm{cc}}L_1^{\mathrm{cc}} + \lambda_{\mathrm{adv}}^{\mathrm{domain}}L_{\mathrm{adv}}^{\mathrm{domain}}+ \\ &\lambda_1^{\mathrm{recon}} L_1^{\mathrm{recon}}
% + \lambda_1^{\mathrm{latent}}L_1^{\mathrm{latent}}+ \lambda_{\mathrm{KL}}L_{\mathrm{KL}}
\end{aligned}
\end{equation}
\vspace{\eqmargin}
%%%%%%%%%%%%%%%%%%%%%%%%%%%%%%%%%%%%%%%%%%%%%%%%%%%%%%%%%%
%%%%%%%%%%%%%%    Experimental Results    %%%%%%%%%%%%%%%%
%%%%%%%%%%%%%%%%%%%%%%%%%%%%%%%%%%%%%%%%%%%%%%%%%%%%%%%%%%
%
\begin{table*}[t]
	\centering
	\caption{\textbf{Summary of the components used in each method.} We desciribe the differences among DRIT, DRIT++, and variants.}
	\begin{tabular}{l c c c}
	\toprule
	Method & mode-seeking & multi-domain & high-resolution \\
	\midrule
	DRIT & - & - & - \\
	DRIT++ (two-domain) & \cmark& - & - \\
	DRIT++ (multi-domain) &\cmark & \cmark& - \\
	DRIT++ (high-resolution) &\cmark& - & \cmark\\
	\bottomrule
	\end{tabular}
	\label{tab:summary}
\end{table*}

%-------------------------------------------------------------
\begin{figure*}[t]
	\centering
     \mpage{0.16}{Input}\hfill\mpage{0.8}{Generated images}
    \includegraphics[width=\linewidth]{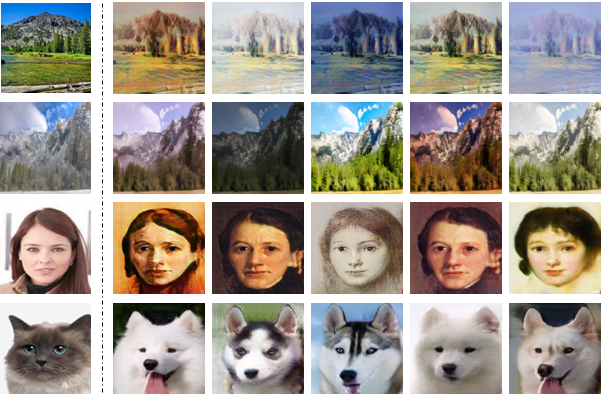}
     
    \caption{\textbf{Sample results.} We show example results produced by our model. The left column shows the input images in the source domain. The other five columns show the output images generated by sampling random vectors in the attribute space. 
    The mappings from top to bottom are: Photo $\rightarrow$ Monet, winter $\rightarrow$ summer, photograph  $\rightarrow$ portrait, and cat $\rightarrow$ dog.
    }
    \label{figure:example}
    %\vspace{\figmargin}
 \end{figure*}
%-------------------------------------------------------------
%-------------------------------------------------------------
\begin{figure*}[t]
    \centering
    \subfloat{%
	    \includegraphics[width=\linewidth]{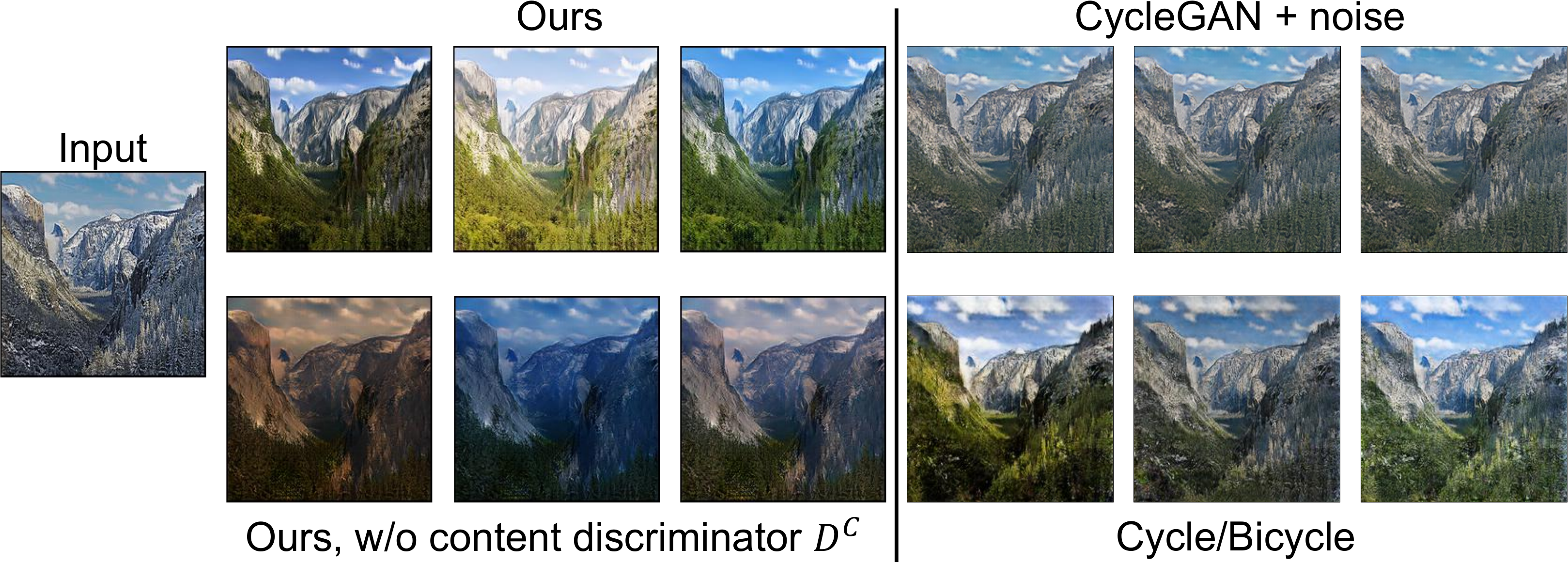}%
	}
	\caption{\textbf{Baseline artifacts.} On the winter $\rightarrow$ summer translation task, our model produces more diverse and realistic samples over baselines.
	}
	\label{figure:diversity}
    %%\vspace{-4mm}
\end{figure*}
%-------------------------------------------------------------
%-------------------------------------------------------------
\begin{figure*}[t]
    \centering
    \subfloat{%
	    \includegraphics[width=\linewidth]{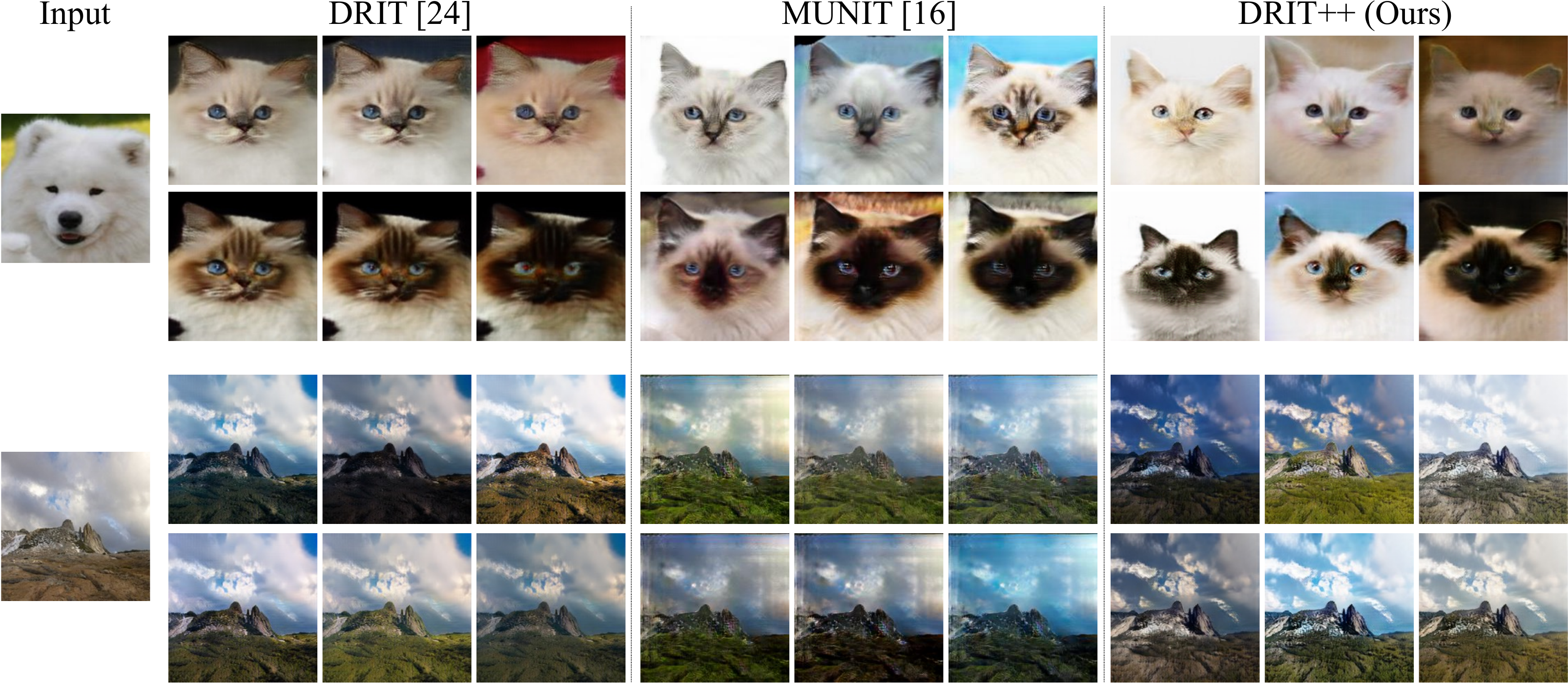}%
	}
	\caption{\textbf{Effectiveness of mode seeking regularization.} 
	Mode seeking regularization helps improve the diversity of translated images while maintaining the visual quality.
	}
	\label{figure:msgan}
    %%\vspace{-4mm}
\end{figure*}
%-------------------------------------------------------------
%-------------------------------------------------------------
\begin{figure*}[t]
	\centering
	\subfloat{%
	    \includegraphics[width=\linewidth]{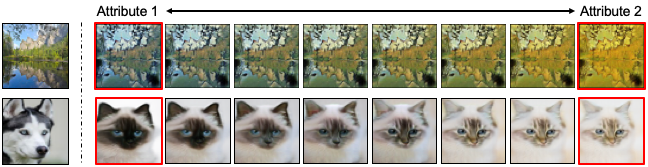}%
	}
	\caption{\textbf{Linear interpolation between two attribute vectors.}  Translation results with linear-interpolated attribute vectors between two attributes (highlighted in red).
	}
	\label{figure:interpolation}
    %\vspace{\figmargin}
\end{figure*}
%-------------------------------------------------------------
%-------------------------------------------------------------
\begin{figure*}[t]
	\centering
	\subfloat[Inter-domain attribute transfer]{%
    	\mpage{0.45}{
    		\mpage{0.28}{Content}\hfill\mpage{0.28}{Attribute}\hfill\mpage{0.3}{Output}
			\includegraphics[width=\linewidth]{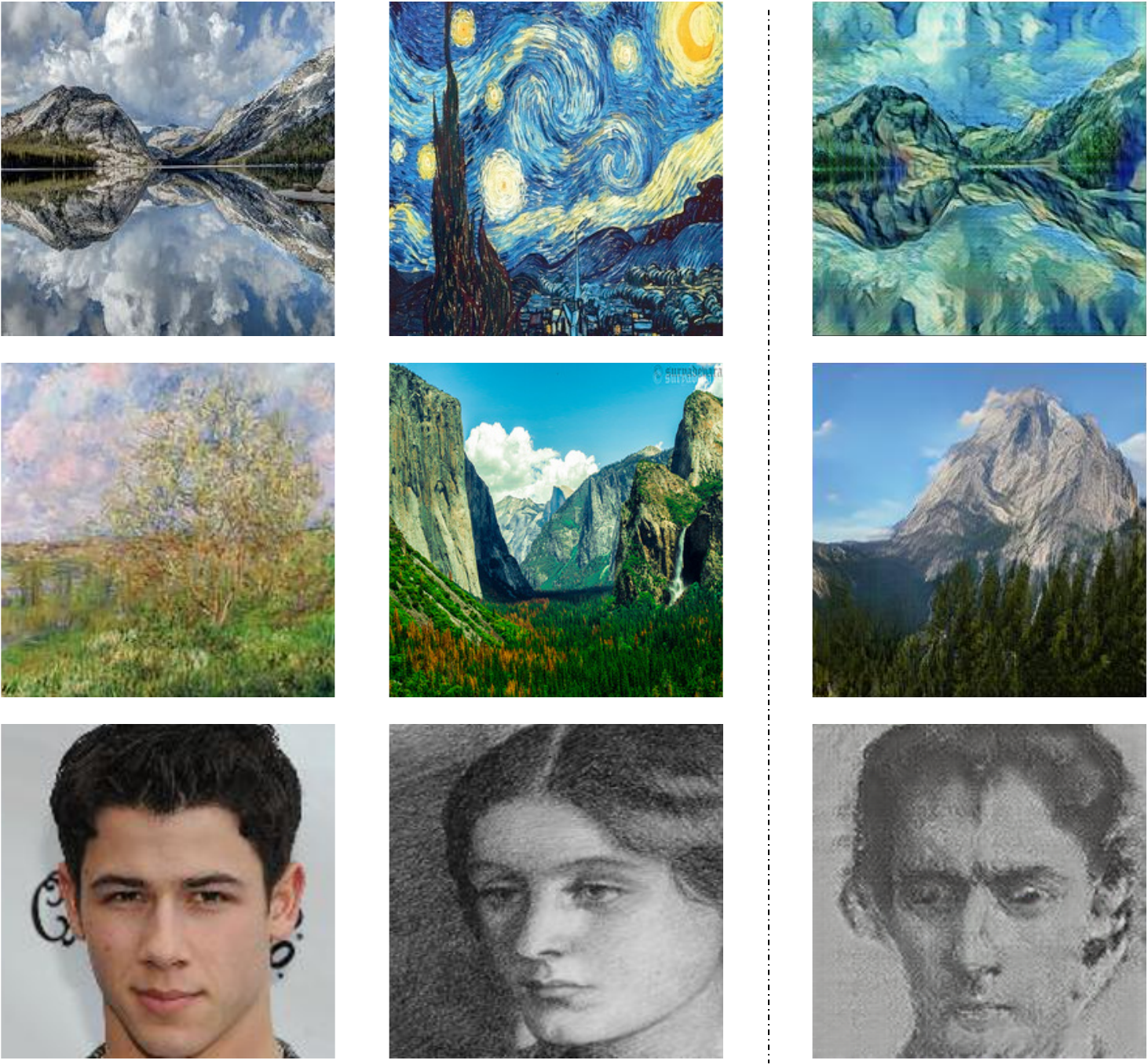}%
        }
     }
  %  }
    \hfill
    \subfloat[Intra-domain attribute transfer]{%
          \mpage{0.45}{
     	\mpage{0.28}{Content}\hfill\mpage{0.28}{Attribute}\hfill\mpage{0.3}{Output}
		\includegraphics[width=\linewidth]{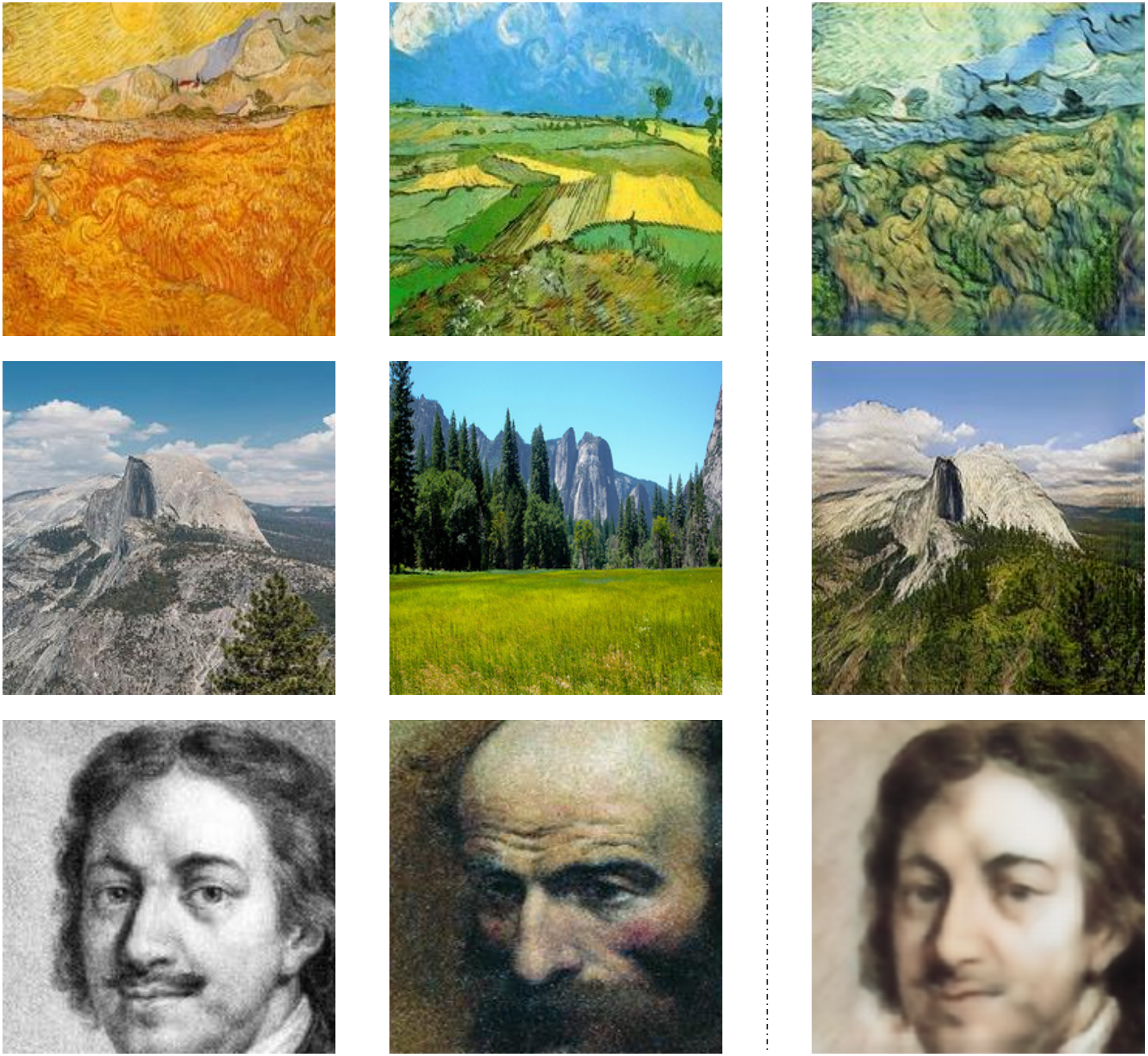}%
	   }
    }
	\caption{\textbf{Attribute transfer.} At test time, in addition to random sampling from the attribute space, we can also perform translation with the query images with the desired attributes. Since the content space is shared across the two domains, we not only can achieve (a) inter-domain, but also (b) intra-domain attribute transfer. Note that we do not explicitly involve intra-domain attribute transfer during training.
	}
	\label{figure:cross}
    %\vspace{\figmargin}
\end{figure*}
%-------------------------------------------------------------

% \vspace{\secmargin}
\section{Experimental Results}
\label{sec:experiments}
% \vspace{\secmargin}
%%\vspace{-1mm}
\Paragraph{Implementation details.}
We implement the proposed model with PyTorch~\cite{paszke2017pytorch}.
We use the input image size of $216 \times 216$ for all of our experiments.
For the content encoder $E^c$, we use an architecture consisting of three convolution layers followed by four residual blocks.
For the attribute encoder $E^a$, we use a CNN architecture with four convolution layers followed by fully-connected layers.
We set the size of the attribute vector to $z^a \in R^8$ for all experiments.
For the generator $G$, we use an architecture consisting of four residual blocks followed by three fractionally strided convolution layers.
%
%For more details of architecture design, please refer to the supplementary material.
%\jiabin{Where is the supplementary material?}

For training, we use the Adam optimizer~\cite{kinga2015adam} with a batch size of $1$, a learning rate of $0.0001$, and exponential decay rates $(\beta_1, \beta_2) = (0.5, 0.999)$.
In all experiments, we set the hyper-parameters as follows: $\lambda^{\mathrm{content}}_{\mathrm{adv}}=1$, $ \lambda_{\mathrm{cc}}=10$, $\lambda^{\mathrm{domain}}_{\mathrm{adv}}=1$, $ \lambda_1^{\mathrm{rec}} =10$,  $\lambda_1^{\mathrm{latent}}=10$, and $\lambda_{\mathrm{KL}}=0.01$. 
We also apply an L1 weight regularization on the content representation with a weight of $0.01$. 
We follow the procedure in DCGAN~\cite{radford2016dcgan} for training the model with adversarial loss.
More results can be found at \url{http://vllab.ucmerced.edu/hylee/DRIT_pp/}. 
The source code and trained models will be made available to the public. 
%
%MH: you can create a webpage with more results, especially those animation in Figure 8-11

\vspace{\paramargin}
\Paragraph{Datasets.}
We evaluate the proposed model on several datasets include Yosemite~\cite{zhu2017cyclegan} (summer and winter scenes), pets (cat and dog) cropped from Google images, artworks~\cite{zhu2017cyclegan} (Monet), and photo-to-portrait cropped from subsets of the WikiArt dataset\footnote{\url{https://www.wikiart.org/}} and the CelebA dataset~\cite{liu2015celeb}. 
%
%We also perform domain adaptation on the classification task with MNIST~\cite{lecun1998MNIST} to MNIST-M~\cite{ganin2016MNISTM}, and on the classification and pose estimation tasks with Synthetic Cropped LineMod to Cropped LineMod~\cite{hinterstoisser2012linemod,wohlhart2015croplinemod}.

%%%\vspace{-3mm}
\vspace{\paramargin}
\Paragraph{Evaluated methods.}
We perform the evaluation on the following algorithms:
%\begin{compactitem}
\begin{itemize}
\item \tb{DRIT++:} The proposed model.
\item \tb{DRIT}~\cite{DRIT}, and \tb{MUNIT}~\cite{huang2018munit}: Multimodal generation frameworks trained with unpaired data.
\item \tb{DRIT w/o $D^c$: } DRIT model without the content discriminator.
\item \tb{Cycle/Bicycle: } We construct a baseline using a combination of CylceGAN and BicycleGAN. Here, we first train CycleGAN on unpaired data to generate corresponding images as \emph{pseudo} image pairs. We then use this pseudo paired data to train BicycleGAN.
\item\tb{CycleGAN}~\cite{zhu2017cyclegan}, and  \tb{BicycleGAN}~\cite{zhu2017bicyclegan}
\end{itemize}
%\end{compactitem}

The proposed DRIT++ method extends the original DRIT method by 1) incorporating mode-seeking regularization for improving sample diversity and 2) generalizing the two-domain model to handle multi-domain image-to-image translation problems.
The DRIT++ (multi-domain)  algorithm is \emph{backward compatible} with the DRIT++ (two-domain) and DRIT methodss with comparable performance (as shown in \subsecref{subsec:quan}).
Thus, the DRIT++ (two-domain) method can be viewed as a special case of the DRIT++ (multi-domain) algorithm. 
The DRIT++ (two-domain) algorithm can improve the visual quality slightly over the DRIT++ (multi-domain) scheme with a category-specific generator and discriminator under the two-domain setting.

%-------------------------------------------------------------
\begin{figure*}[t]
	\centering
	\subfloat{%
	    \includegraphics[width=0.95\linewidth]{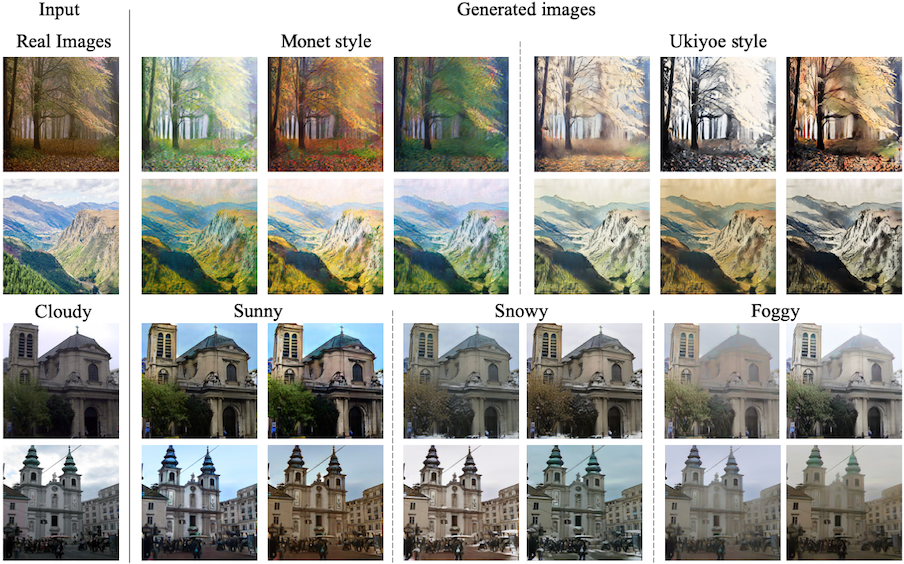}%
	}
	\caption{\textbf{Multi-domain I2I.}
	We show example results of our model on the multi-domain I2I task.
	We demonstrate the translation among real images and two artistic styles (Monet and Ukiyoe), and the translation among different weather conditions (sunny, cloudy, snowy, and foggy).
	}
	\label{figure:multi}
    %\vspace{\figmargin}
\end{figure*}
%-------------------------------------------------------------

%-------------------------------------------------------------
\begin{figure*}[t]
	\centering
    \includegraphics[width=\linewidth]{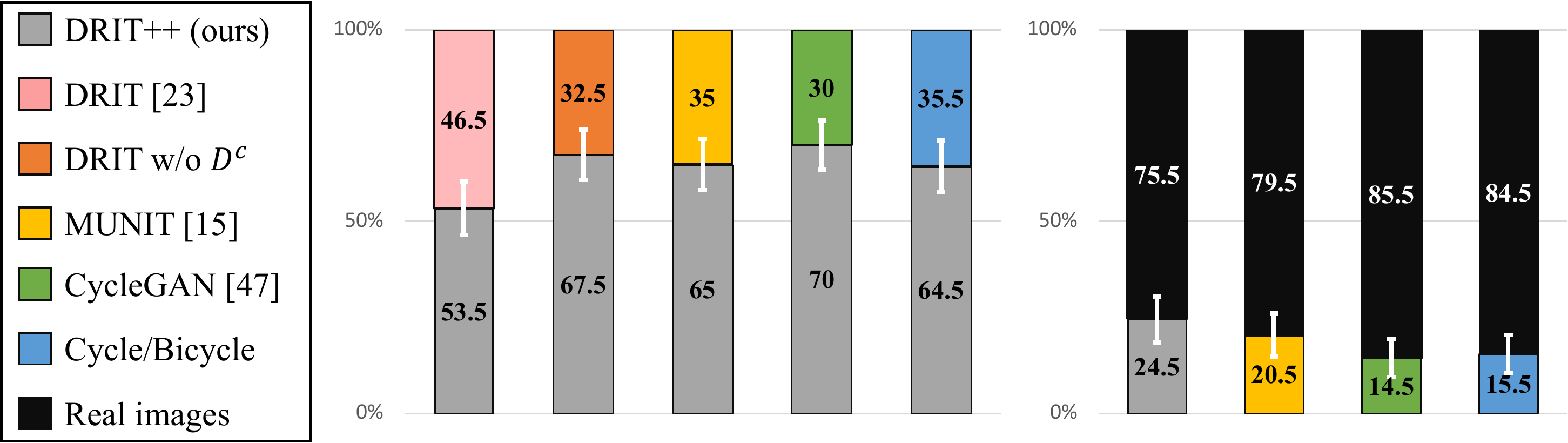}
    \caption{\textbf{Realism of synthesized images.} 
    We conduct a user study to ask subjects to select results that are \emph{more realistic} through pairwise comparisons.
    The number indicates the percentage of preference for that comparison pair. 
    We use the winter $\rightarrow$ summer and the cat $\rightarrow$ dog translation for this experiment.}
    \label{figure:realism}
    %%\vspace{-5mm}
 \end{figure*}
%-------------------------------------------------------------

\subsection{Qualitative Evaluation}
%% \vspace{\subsecmargin}
%%\vspace{-1mm}
\Paragraph{Diversity.} We first compare the proposed model with other methods in \figref{example}.
In \figref{diversity}, demonstrate the visual artifacts of images generated by baseline methods. 
Both our model without $D^c$ and Cycle/Bicycle can generate diverse results. 
However, the results contain clearly visible artifacts. 
Without the content discriminator, our model fails to capture domain-specific details (\eg the color of tree and sky).
Therefore, the variations of synthesized images lie in global color differences.
As the Cycle/Bicycle methods are trained on pseudo paired data generated by CycleGAN, the quality of the pseudo paired data is not high.
As a result, the generated images contain limited diversity. 

To better analyze the learned domain-specific attribute space, we perform linear interpolation between two given attributes and generate the corresponding images as shown in \figref{interpolation}.
The interpolation results validate the continuity in the attribute space and show that our model can generalize in the distribution, rather than simply retain visual information.

\vspace{\paramargin}
\Paragraph{Mode seeking regularization.}
We demonstrate the effectiveness of the mode seeking regularization term in \figref{msgan}.
The mode seeking regularization term substantially alleviates the mode collapse issue in DRIT~\cite{DRIT}, particularly in the challenging shape-variation translation (\ie dog-to-cat translation).

\vspace{\paramargin}
\Paragraph{Attribute transfer.} 
We demonstrate the results of the attribute transfer in \figref{cross}.
By disentangling content and attribute representations, we are able to perform attribute transfer from images of desired attributes, as illustrated in \figref{architecture}(c).
Furthermore, since the content space is shared between two domains, we can generate images conditioned on content features encoded from either domain.
Thus our model can achieve not only inter-domain but also intra-domain attribute transfer.
Note that intra-domain attribute transfer is not explicitly involved in the training process.

\vspace{\paramargin}
\Paragraph{Multi-domain I2I.} 
%MH: Why do you refer to Figure 5 here?
%\figref{mdmm} shows the results of applying the proposed method on the multi-domain I2I.
\figref{multi} shows the results of applying the proposed method on the multi-domain I2I.
We perform translation among three domains (real images and two artistic styles) and four domains (different weather conditions).
Using one single generator, the proposed model is able to perform diverse translation among multiple domains.

\subsection{Quantitative Evaluation}
\label{subsec:quan}
 % \vspace{\subsecmargin}
 
\Paragraph{Metrics}
We conduct quantitative evaluations using the following metrics:
%\begin{compactitem}
\begin{itemize}
\item\tb{FID. }
To evaluate the quality of the generated images, we use the FID~\cite{heusel2017gans} metric to measure the distance between the generated distribution and the real one through features extracted by Inception Network~\cite{szegedy2015going}.
Lower FID values indicate better quality of the generated images.

\item\tb{LPIPS.}
To evaluate diversity, we employ LPIPS~\cite{zhang2018perceptual} metric to  measure the average feature distances between generated samples.
Higher LPIPS scores indicate better diversity among the generated images.

\item\tb{JSD and NDB. }
To measure the similarity between the distribution between real images and generated one, we
adopt two bin-based metrics, JSD and NDB~\cite{richardson2018NDB}.
These metrics evaluate the extent of mode missing of generative models.
Similar to~\cite{richardson2018NDB}, we first cluster the training samples using K-means into different bins.
These bins can be viewed as modes of the real data distribution.
We then assign each generated sample to the bin of its nearest neighbor.
We compute the bin-proportions of the training samples and the synthesized samples to evaluate the difference between the generated distribution and the real data distribution.
%\hungyu{testing samples? or generated/synthesized samples? of what models?}
%
The NDB and JSD metrics of the bin-proportion are then computed to measure the level of mode collapse.  
Lower NDB and JSD scores mean the generated data distribution approaches the real data distribution better by fitting more modes.
More discussions on these metrics can be found in~\cite{richardson2018NDB}.

\item\tb{User preference.}
For evaluating realism of synthesized images, we conduct a user study using pairwise comparison.
Given a pair of images sampled from real images and translated images generated from various methods, each subject needs to answer the question  ``Which image is more realistic?''
%\end{compactitem}
\end{itemize}

\Paragraph{Realism \vs diversity.} 
We conduct the experiment using winter $\rightarrow$ summer and cat $\rightarrow$ dog translation with the Yosemite and pets datasets, respectively.
\tabref{drit}, \tabref{ablation}, and \figref{realism} present the quantitative comparisons with other methods as well as baseline methods.
In \tabref{drit}, the DRIT++ method performs well on all metrics.
The DRIT++ method generates images that are not only realistic, but also diverse and close to the original data distribution.
\tabref{ablation} validates the effectiveness of the content discriminator, latent regression loss, and mode-seeking regularization in the proposed algorithm. 
\figref{realism} shows the results of user study.
The DRIT++ algorithm performs favorably against the state-of-the-art approaches as well as baseline methods.

\begin{figure}[h]
    \centering
    \includegraphics[width=\linewidth]{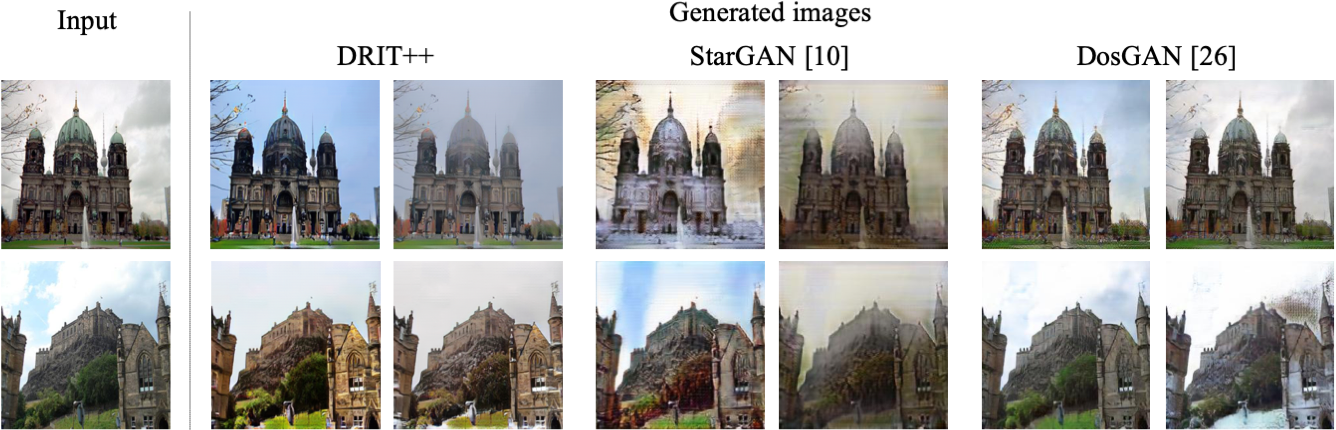}
    \caption{\textbf{Comparisons of different multi-domain translation model on the weather dataset.}}
    \label{figure:MD_compare}
\end{figure}

\Paragraph{Multi-domain translation}
We compare the performance of DRIT++,  StarGAN~\cite{StarGAN2018}, and DosGAN~\cite{lin2019explore} in terms of realism on the weather dataset.
For each trial, We translate 1000 testing images to one of four domains and measure the visual quality (in terms of FID) and diversity (using the LPIPS metric).
We report the averaged results of 5 trials.
\tabref{MD_compare} shows that the disentangled representations by our method not only enable diverse translation, but also improve the quality of generated images.
\figref{MD_compare} presents qualitative results by the evaluated methods.

\Paragraph{Multi-domain model on two-domain translation}
Two-domain translation is a special case of multi-domain translation problems.
%
%MH2: why do you hand code the table numbers? It is fine that you copy sentence from the response to comments file, but you do not need to check the reference numbers.
%We conduct an experiment under the same setting as the quantitative experiment in Table 1 and Table 2 in the paper. 
We conduct an experiment under the same settings described in 
Table \ref{tab:drit} and \ref{tab:MDMM}.
As shown in \tabref{MDMM}, our multi-domain model performs well in all metrics against the two-domain translation model that consists of the domain-specific generator and discriminator.

\begin{figure}[h]
    \centering
    \includegraphics[width=0.6\linewidth]{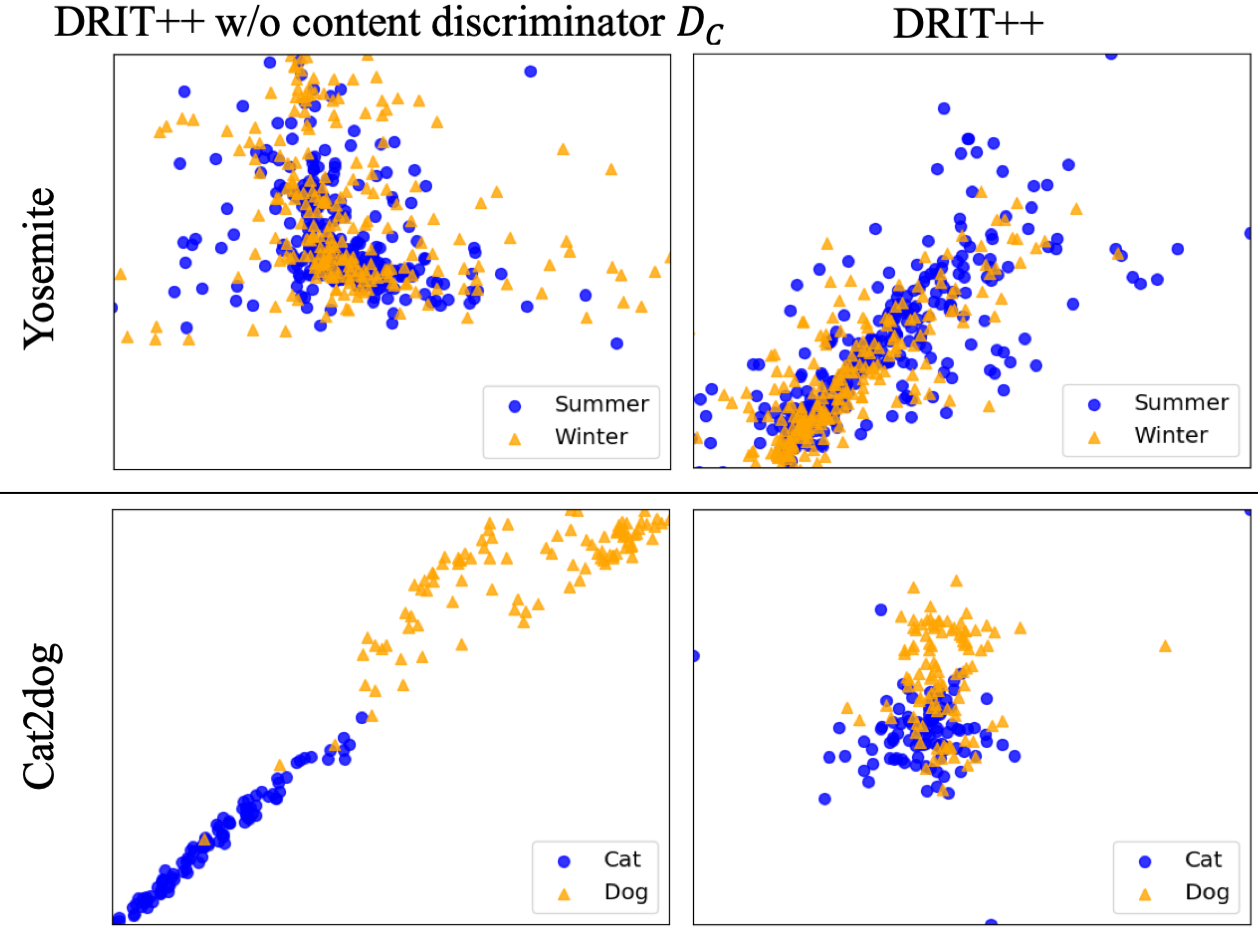}
    \caption{\textbf{Visualization of the latent content representations of two domains using t-SNE.}
    Each data point is a content representation encoded from an image of that domain.
    }
    \label{figure:DC}
\end{figure}

\Paragraph{Ablation study on the content discriminator}
In practice, the content discriminator helps align distributions of the latent content representations of two domains.
We conduct experiments on both cat2dog and the Yosemite datasets to illustrate this. 
The distance between the means of the content representations from two domains is measured by:
%
%MH: if you treat equations as parts of the text and end them with comma or dots, then do that. Be consistent. 
\begin{equation}
\label{eq:dis}
    D = \left\|\frac{1}{N_{A}^\mathrm{test}}\sum_{i=1}^{N_A^\mathrm{test}} f_A^\mathrm{content} - \frac{1}{N_{B}^\mathrm{test}}\sum_{i=1}^{N_B^\mathrm{test}} f_B^\mathrm{content}\right\|_1^1.
\end{equation}
\tabref{DC} shows the quantitative results.
Furthermore, \figref{DC} visualizes the distributions of the latent content representations from two domains using t-SNE.
The distance \eqnref{dis} between the content representations of the two domains is much smaller with the help of the content discriminator.

%%%%%%%%%%%%%%%%%%%%%%%%%%%%%%%%%%%%%%%%%%%%%%%%%%%%%%%%%%%%%%%%%%%%%%
\begin{table*}[t]
	\centering
	\caption{\textbf{Quantitative results of the Yosemite (Summer$\rightleftharpoons$Winter) and the Cat$\rightleftharpoons$Dog dataset.}}
	
	%%\vspace{-1mm}
	\begin{tabular}{@{}cccccc} 
	    \toprule
		Datasets & \multicolumn{4}{c}{Winter $\rightarrow$ Summer}
		\\  \cmidrule(lr){2-5} 
		&   Cycle/Bicycle &     DRIT         &       MUNIT       &     DRIT++ \\
		FID $\downarrow$ &  $67.04\pm{0.60}$ &$41.34\pm{0.20}$  & $57.09\pm{0.37}$          &$\mathbf{41.02\pm{0.24}}$ \\ 
		NDB$\downarrow$  & $9.36\pm{0.69}$              &  $9.38\pm{0.74}$& $9.53\pm{0.64}$                 & $\mathbf{9.22\pm{0.97}}$ \\
		JSD$\downarrow$  & $0.290\pm{0.086}$               &$0.304\pm{0.075}$ & $0.293\pm{0.062}$                &$\mathbf{0.222\pm{0.070} }$\\
		LPIPS$\uparrow$  & $0.0974\pm{0.0003}$              & $0.0965\pm{0.0004}$& $0.1136\pm{0.0008}$                 & $\mathbf{0.1183\pm{0.0007}}$
		\\ \midrule
		Datasets&  \multicolumn{4}{c}{Cat $\rightarrow$ Dog}
		\\ \cmidrule(lr){2-5} 
	    & Cycle/Bicycle   &     DRIT         &       MUNIT       & DRIT++ \\
		FID$\downarrow$ & $54.008\pm{1.590}$    &$24.306\pm{0.329}$   &       $22.127\pm{0.712}$ &$\mathbf{17.253\pm{0.648}}$\\ 
		NDB$\downarrow$ & $9.23\pm{0.84}$               &  $8.16\pm{1.60}$&        $8.21\pm{1.17}$        &$\mathbf{7.57\pm{1.25}}$   \\
		JSD$\downarrow$ & $0.262\pm{0.072}$                &$0.075\pm{0.046}$ &        $0.132\pm{0.066}$        &$\mathbf{0.041\pm{0.014}}$ \\
		LPIPS$\uparrow$ & $0.147\pm{0.001}$         &$0.245\pm{0.002}$ &$0.244\pm{0.002}$        &$\mathbf{0.280\pm{0.002}}$\\
		\bottomrule 
		
	\end{tabular}
	\label{tab:drit}
	%%\vspace{-3mm}
\end{table*}
%%%%%%%%%%%%%%%%%%%%%%%%%%%%%%%%%%%%%%%%%%%%%%%%%%%%%%%%%%%%%%%%%%%%%%

\begin{table*}[h]
	\centering
	\caption{\textbf{Quantitative results of DRIT++ (multi-domain) on the Yosemite (Summer$\rightleftharpoons$Winter) dataset.}
	%\jiabin{Be consistent. multi-domain DRIT++ -> DRIT++ (multi-domain). This reduces mental load for people parsing your paper.}
	}
	\begin{tabular}{c c c}
	\toprule
	      & DRIT++ (two-domain) &  DRIT++ (multi-domain)  \\
	\midrule
	FID $\downarrow$& $\mathbf{41.02\pm{0.24}}$ & $44.86\pm0.33$ \\
	NDB $\downarrow$& $\mathbf{9.22\pm{0.97}}$ & $9.20\pm0.88$\\
	JSD $\downarrow$& $\mathbf{0.222\pm{0.070} }$& $0.254 \pm 0.051$\\
	LPIPS$\uparrow$ & $0.1183\pm{0.0007}$&$\mathbf{0.1204\pm{0.0004}}$ \\
	\bottomrule
	\end{tabular}
	\label{tab:MDMM}
\end{table*}
%%%%%%%%%%%%%%%%%%%%%%%%%%%%%%%%%%%%%%%%%%%%%%%%%%%%%%%%%%%%%%%%%%%%%%

\begin{table*}[h]
	\centering
	\caption{\textbf{Average distance between latent content representations of two domains.}}
	\begin{tabular}{c c c}
	\toprule
	Dataset    & DRIT++ & DRIT++ w/o content discriminator $D_c$ \\
	\midrule
	cat2dog & \textbf{10.45}  & 55.45 \\
	Yosemite & \textbf{31.56} & 58.69 \\
	\bottomrule
	\end{tabular}
	\label{tab:DC}
\end{table*}

%%%%%%%%%%%%%%%%%%%%%%%%%%%%%%%%%%%%%%%%%%%%%%%%%%%%%%%%%%%%%%%%%%%%%%%%%%%%
\begin{table*}[t]
	\centering
	\caption{\textbf{Ablation study.} We demonstrate the effect of content discriminator, latent regression loss, and mode-seeking regularization in the proposed algorithm. }
	\begin{tabular}{@{}cccccc} 
	    \toprule
		                 &    DRIT w/o $D^c$  &   DRIT w/o KL &     DRIT w/o $L_1^{\mathrm{latent}}$        &       DRIT       &     DRIT++ \\
		FID $\downarrow$ &  $46.92\pm{0.35}$ &    $\mathbf{40.08\pm{0.33}}$          &$53.12\pm{0.16}$                        &$41.34\pm{0.20}$  &$41.02\pm{0.24}$ \\ 
		NDB$\downarrow$  &  $9.36\pm{0.72}$  & $9.47\pm{0.70}$              &$9.97\pm{0.17}$                         &  $9.38\pm{0.74}$                 & $\mathbf{9.22\pm{0.97}}$ \\
		JSD$\downarrow$  & $0.277\pm{0.077}$ & $0.289\pm{0.066}$             &$0.494\pm{0.045}$                        &$0.304\pm{0.075}$                &$\mathbf{0.222\pm{0.070} }$ \\
		LPIPS$\uparrow$  &   $0.0954\pm{0.0006}$              & $0.0957\pm{0.0007}$             &        $0.0158\pm{0.0003}$         & $0.0965\pm{0.0004}$& $\mathbf{0.1183\pm{0.0007}}$\\
		\bottomrule
    \end{tabular}
	\label{tab:ablation}
\end{table*}
%%%%%%%%%%%%%%%%%%%%%%%%%%%%%%%%%%%%%%%%%%%%%%%%%%%%%%%%%%%%%%%%%%%%%%%%%%%%
\begin{table*}[h]
	\centering
	\caption{\textbf{Multi-domain translation comparison.} We compare the visual quality and diversity of DRIT++ (multi-domian) with two multi-domain translation model on the weather dataset. The results are averaged after 5 trials. 
	StarGAN gets highest score on LPIPS due to its lower visual quality.
	}
	\begin{tabular}{c c c c}
	\toprule
	      & DRIT++ & StarGAN & DosGAN  \\
	\midrule 
	FID$\downarrow$ &  $\mathbf{61.51\pm{3.11}}$ & $82.38\pm{3.91}$ &$67.98\pm{2.38}$ \\
	LPIPS$\uparrow$ & $0.676\pm{0.008}$ & $\mathbf{0.692\pm{0.010}}$ & $0.650\pm{0.005}$\\
	\bottomrule
	\end{tabular}
	\label{tab:MD_compare}
\end{table*}

%% ---- high resolution ---
%-------------------------------------------------------------
\begin{figure}
	\centering
	\subfloat{%
	    \includegraphics[width=0.95\linewidth]{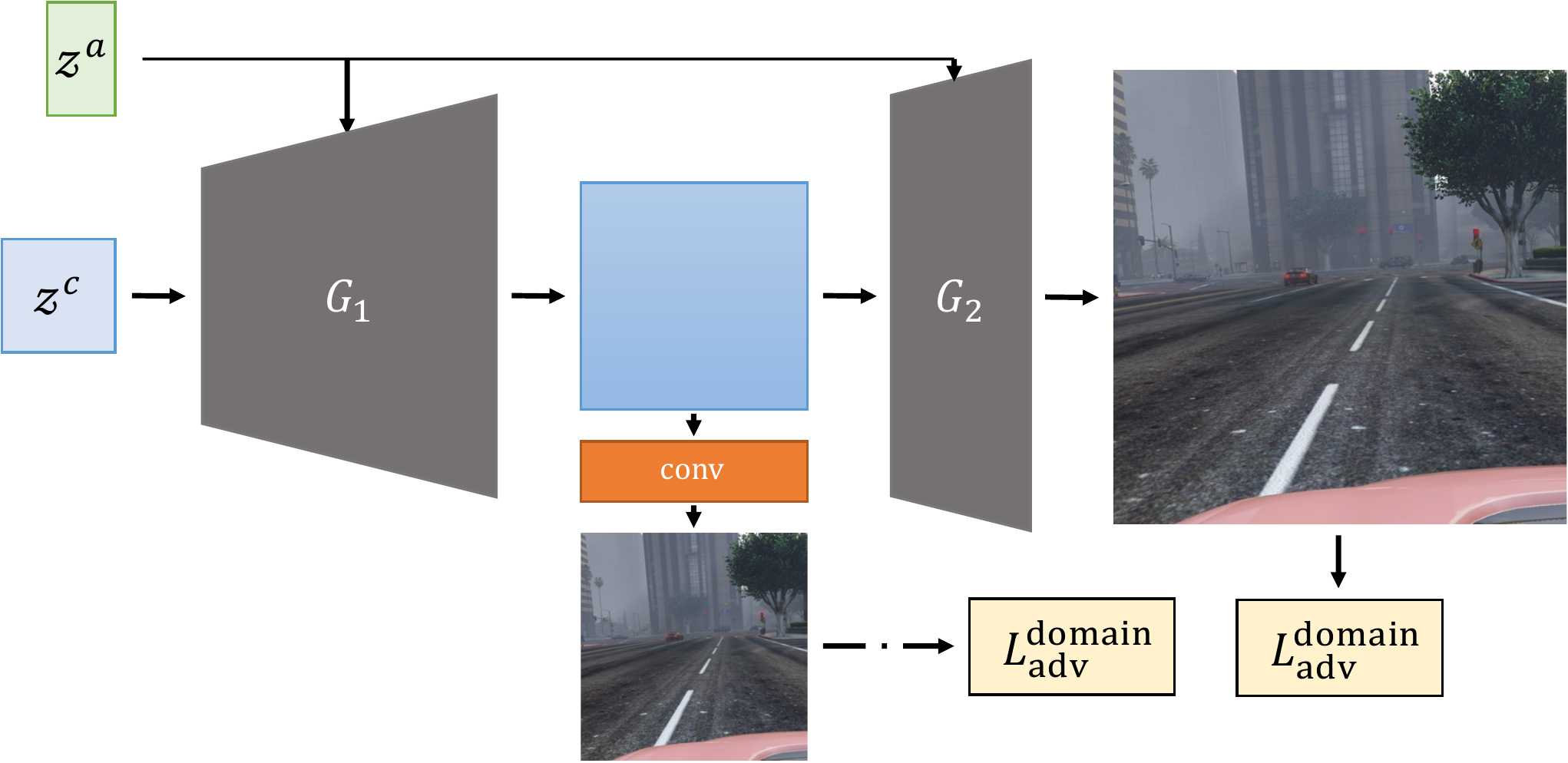}%
	}
	\caption{\textbf{Multi-scale generator-discriminator.} To enhance the quality of generated high-resolution images, we adopt a multi-scale generator-discriminator architecture. We generate low-resolution images from the intermediate features of the generator. An additional adversarial domain loss is applied on the low-resolution images.}
	\label{figure:architecture_hr}
%   %\vspace{-5mm}
\end{figure}

%HERE
%-------------------------------------------------------------
%%\vspace{-1mm}
% \vspace{\secmargin}

\subsection{High Resolution I2I}
\label{subsec:highresolution}
% \vspace{\secmargin}

\begin{figure*}[t]
	\centering
	\subfloat{%
	    \includegraphics[width=\linewidth]{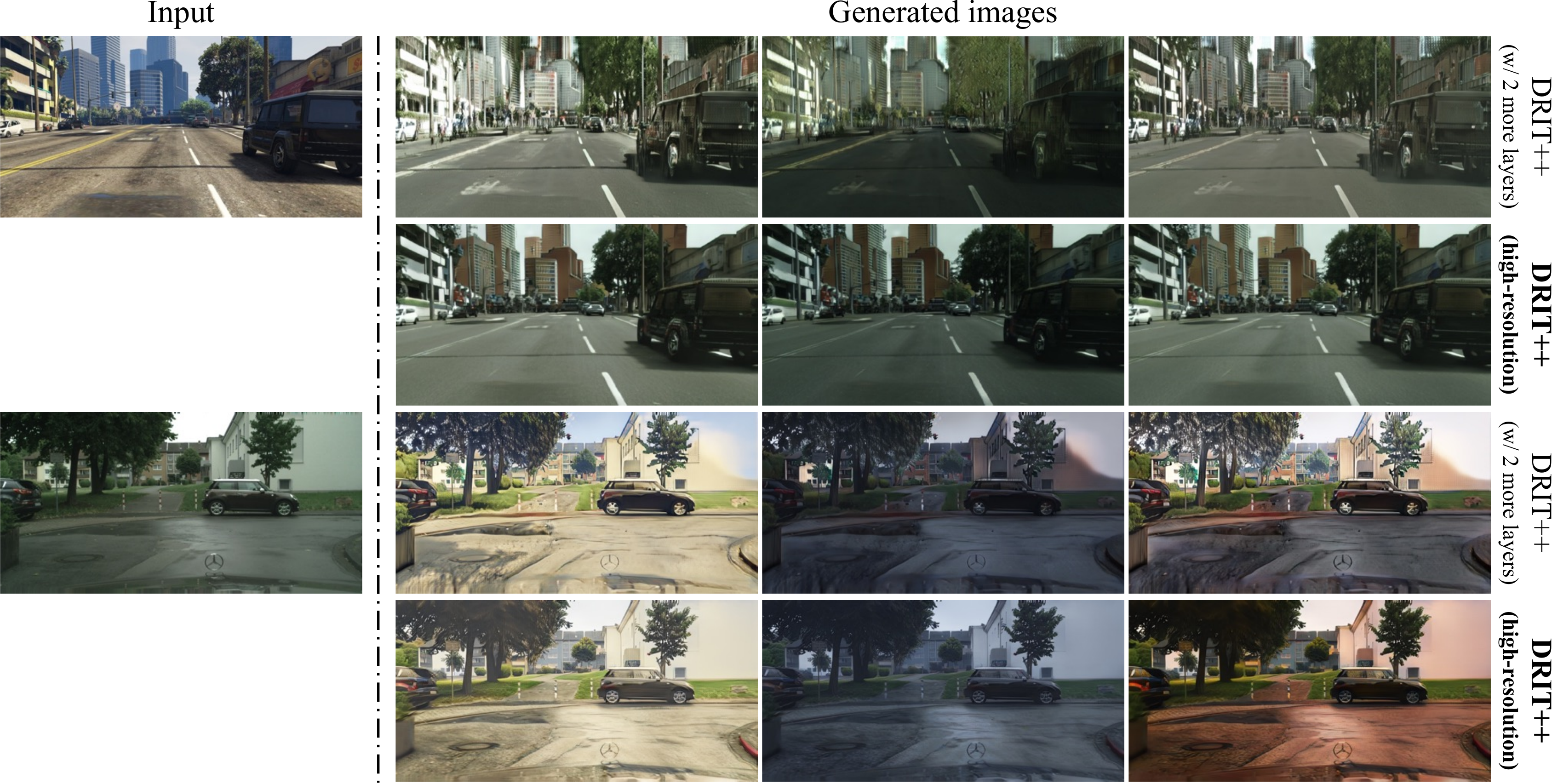}%
	}
	\caption{\textbf{High-resolution translations.} We show sample results produced by our model with multi-scale generator-discriminator architecture. The mappings from top to bottom are: GTA $\rightarrow$ Cityscape, Cityscape $\rightarrow$ GTA.}
	\label{figure:result_hd}
    %\vspace{-5mm}
\end{figure*}

We demonstrate that the proposed scheme can be applied to the translation tasks with high-resolution images.
We perform image translation on the street scene (GTA~\cite{Richter_2016_ECCV} $\leftrightarrow$ Cityscape~\cite{cordts2016cityscapes}) dataset.
The size of the input image is $720\times360$ pixels.
During the training, we randomly crop the image to the size of $340\times340$ for memory efficiency consideration.
To enhance the quality of the generated high-resolution images, we adopt a multi-scale generator-discriminator structure similar to the StackGAN~\cite{zhang2017stackgan++} scheme.
As shown in \figref{architecture_hr}, we extract the intermediate feature of the generator and pass through a convolutional layer to generate low-resolution images.
We utilize an additional discriminator which takes low-resolution images as input.
This discriminator enforces the first few layers of the generator to capture the distribution of low-level variations such as colors and image structures.
We find such multi-scale generator-discriminator structure facilitate the training and yields more realistic images on high-resolution translation task.
To validate the effectiveness of the multi-scale architecture, we show the comparison between (1) adding two more layers to generators and (2) using the multi-scale generator-discriminator architecture in \tabref{MS_compare} and \figref{result_hd}.
We report the FID and LPIPS scores of the generated images by the two methods on the GTA5~$\rightarrow$~Cityscape translation task.
As shown in \tabref{MS_compare}, using the multi-scale architecture we can generate more photo-realistic images on the translation task with high-resolution images.

%----------------------------------------------
\begin{table}[h]\tiny
	\centering
	\caption{\textbf{Ablation study on multi-scale generator-discriminator architecture.} We improvement using two more layers in the multi-scale architecture.}
	\begin{tabular}{c c c}
	\toprule
	& DRIT++ w/ 2 more layers & DRIT++ (high-resolution)  \\
	\midrule 
	FID $\downarrow$ & $37.19 \pm 0.21$ & $\mathbf{28.62 \pm 0.38}$ \\
	LPIPS $\uparrow$ & $0.616 \pm 0.038$ & $\mathbf{0.621 \pm 0.004 }$\\
	\bottomrule
	\end{tabular}
	\label{tab:MS_compare}
\end{table}
%-------------------------------------------------------------
%-------------------------------------------------------------

%-------------------------------------------------------------
\begin{figure*}[t]
	\centering
	\subfloat[Summer $\rightarrow$ Winter]{
    \includegraphics[width=0.4\linewidth]{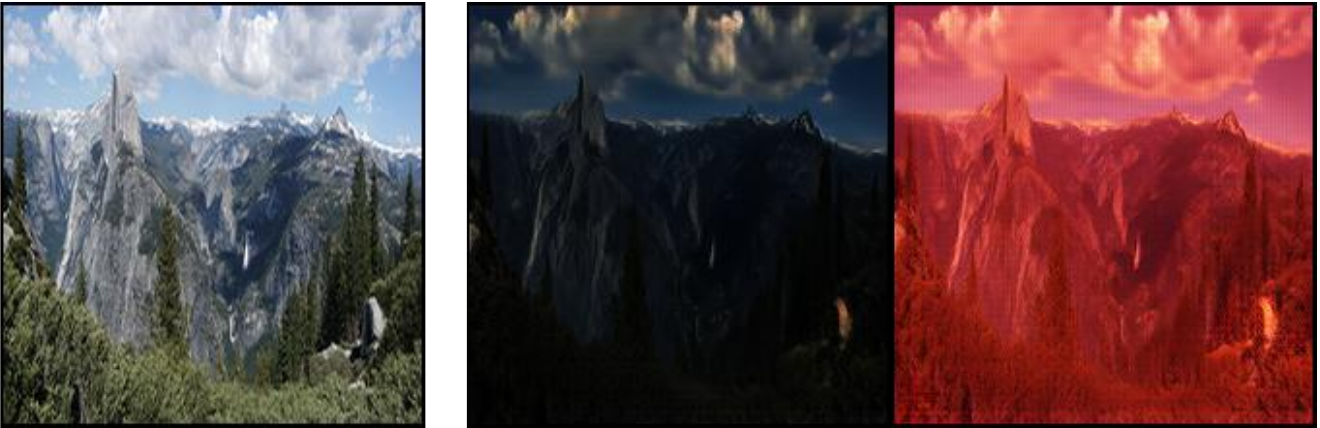}
	}
	\subfloat[van Gogh $\rightarrow$ Monet]{
    \includegraphics[width=0.4\linewidth]{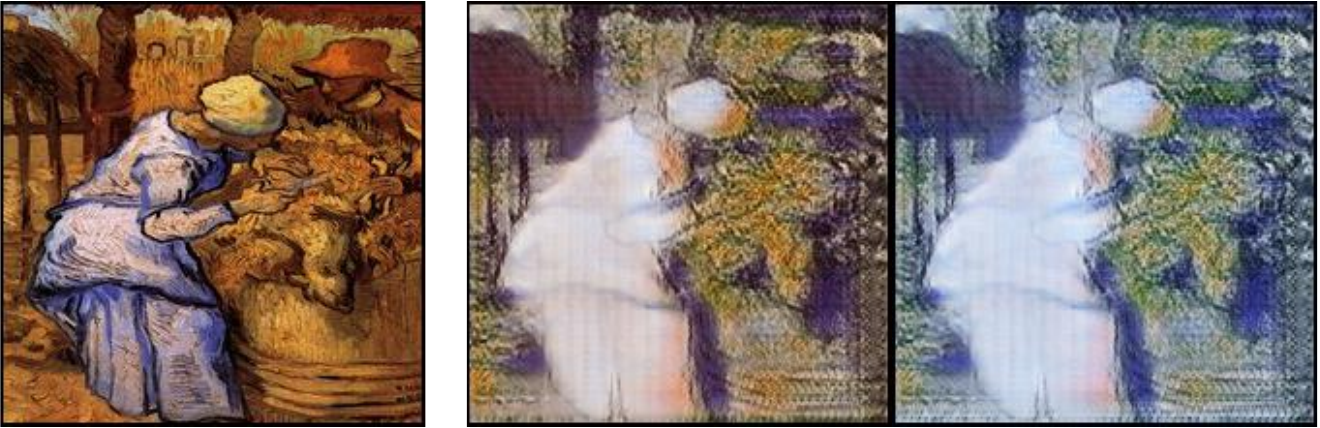}
	}
    \vspace{\subfigmargin}
    %\vspace{-3mm}
   
	\caption{\textbf{Failure examples.} Typical cases: (a) Attribute space not fully exploited. (b) Distribution characteristic difference. }
	\label{figure:failure}
    %%\vspace{-5mm}
\end{figure*}
%-------------------------------------------------------------
%%\vspace{-1mm}
% \vspace{\secmargin}
\subsection{Limitations}
\label{subsec:limitation}
% \vspace{\secmargin}
%
The performance of the proposed algorithm is limited in several aspects. 
First, due to the limited amount of training data, the attribute space is not fully exploited.
Our I2I translation fails when the sampled attribute vectors locate in under-sampled space, see \figref{failure}(a).
Second, it remains difficult when the domain characteristics differ significantly.
For example, \figref{failure}(b) shows a failure case on the human figure due to the lack of human-related portraits in Monet collections.
Third, we use multiple encoders and decoders for the cross-cycle consistency during training, which requires large memory usage.
The memory usage limits the application on high-resolution image-to-image translation.

%%\vspace{-1mm}
\section{Conclusions}
\label{sec:conclusion}
% \vspace{\secmargin}
In this paper, we present a novel disentangled representation framework for diverse image-to-image translation with unpaired data.
we propose to disentangle the latent space to a content space that encodes common information between domains, and a domain-specific attribute space that can model the diverse variations given the same content.
We apply a content discriminator to facilitate the representation disentanglement.
We propose a cross-cycle consistency loss for cyclic reconstruction to train in the absence of paired data.
Qualitative and quantitative results show that the proposed model produces realistic and diverse images.
We also apply the proposed algorithm to domain adaptation and achieve competitive performance compared to the state-of-the-art methods.

\section*{Acknowledgements}
\vspace{-3mm}
This work is supported in part by the NSF CAREER Grant \#1149783, the NSF Grant \#1755785, and gifts from Verisk, Adobe and Google.

%\begin{acknowledgements}
%\end{acknowledgements}

\bibliographystyle{spmpsci}      % mathematics and physical sciences
\bibliography{ijcv19}   % name your BibTeX data base

%\begin{thebibliography}{}
% etc
%\end{thebibliography}

\end{document}